\documentclass[journal]{IEEEtran}
\usepackage{cite}
\usepackage{amsmath}
\usepackage[T1]{fontenc}
\usepackage[latin9]{inputenc}
\usepackage{units}
\usepackage{textcomp}
\usepackage{stmaryrd}
\usepackage[pdftex]{graphicx}
\usepackage[caption=false,font=footnotesize]{subfig}
\usepackage[english]{babel}
\usepackage{stfloats}
\usepackage{array}
\usepackage{color}
\usepackage{url}
\usepackage[ruled]{algorithm2e}
\usepackage{amssymb}
\usepackage{footnote}
\usepackage{amsmath,bm}
\usepackage{amsmath, amssymb}
\usepackage{threeparttable}
\usepackage{ulem}
\usepackage{graphicx}
\usepackage{array}
\usepackage{booktabs}
\usepackage{multirow}
\usepackage{makecell}
\usepackage[pagebackref=false,breaklinks=true,colorlinks,bookmarks=false,unicode=true]{hyperref}

\usepackage{babel}
\makesavenoteenv{table}

\providecommand{\tabularnewline}{\\}

\newcommand{\nickname}{VISO}
\newcommand{\nicknamemethod}{MMB}
\newcommand{\qy}[1]{\textcolor{black}{#1}}

\ifCLASSINFOpdf
\else
\fi

\begin{document}
\title{Detecting and Tracking Small and Dense Moving Objects in Satellite Videos: A Benchmark}

\author{Qian Yin\textsuperscript{$\ast$}, Qingyong Hu\textsuperscript{$\ast$}, Hao Liu, Feng Zhang, Yingqian Wang, Zaiping Lin, Wei An, Yulan Guo
\thanks{Q.~Yin, F.~Zhang, Y.~Wang, Z.~Lin, W.~An, and Y.~Guo are with the College of Electronic Science and Technology, National University of Defense Technology, P. R. China. Q. Hu is with the Department of Computer Science, University of Oxford, UK. H. Liu and Y. Guo are with the School of Electronics and Communication Engineering, the Shenzhen Campus of Sun Yat-sen University, P. R. China. Emails: yinqian18@nudt.edu.cn, guoyulan@sysu.edu.cn. (Corresponding author: Yulan~Guo).

*Q.~Yin and Q. Hu have equal contribution to this work and are co-first authors.}}

\markboth{Journal of \LaTeX\ Class Files,~Vol.~14, No.~8, August~2015}%
{Shell \MakeLowercase{\textit{et al.}}: Bare Demo of IEEEtran.cls for IEEE Journals}

\maketitle

\begin{abstract}

Satellite video cameras can provide continuous observation for a large-scale area, which is important for many remote sensing applications. However, achieving moving object detection and tracking in satellite videos remains challenging due to the insufficient appearance information of objects and lack of high-quality datasets. In this paper, we first build a large-scale satellite video dataset with rich annotations for the task of moving object detection and tracking. \qy{This dataset is collected by the Jilin-1 satellite constellation and composed of 47 high-quality videos with 1,646,038 instances of interest for object detection and 3,711 trajectories for object tracking.} We then introduce a motion modeling baseline to improve the detection rate and reduce false alarms based on accumulative multi-frame differencing and robust matrix completion. Finally, we establish the first public benchmark for moving object detection and tracking in satellite videos, and extensively evaluate the performance of several representative approaches on our dataset. Comprehensive experimental analyses and insightful conclusions are also provided. \qy{The dataset is available at \url{https://github.com/QingyongHu/VISO}.}


\end{abstract}


\IEEEpeerreviewmaketitle

\section{Introduction}

\begin{figure}[t]
\centering
\includegraphics[width=8.5cm]{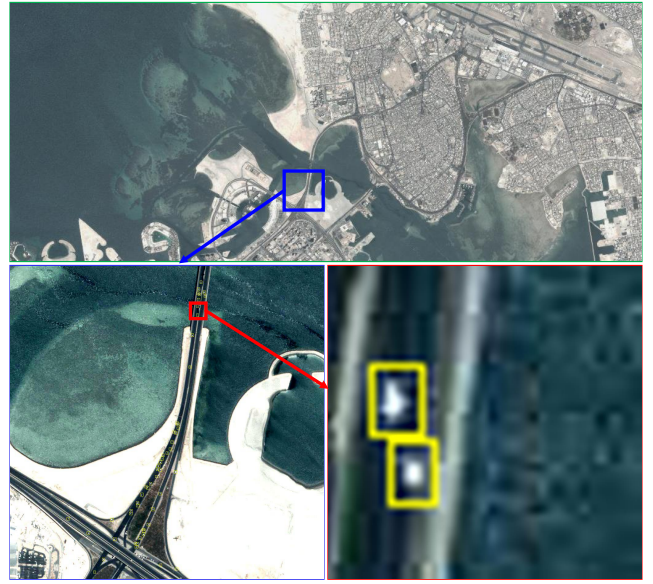}
\caption{\qy{Example frame of Muharraq, Bahrain from the proposed \nickname{} satellite video dataset. Note that, the area with a blue bounding box on the left is enlarged by 50 times to obtain the bottom left image. Then, the area with a red bounding box is further zoomed in to obtain the image shown in the bottom right. Finally, the yellow bounding box represents the ground-truth annotation of the object of interest.}}\label{thumbnail}
\end{figure}

\IEEEPARstart{M}{oving}
\qy{object detection and tracking in video sequences plays an important role in video surveillance \cite{videosurveillance}, digital city \cite{digital1}, and intelligent traffic management
\cite{intelligent}. Major platforms for moving object detection and tracking include closed-circuit televisions (CCTVs) \cite{CCTV}, aircrafts \cite{aircraft1,aircraft2}, and unmanned aerial vehicles (UAVs) \cite{uav1}. Videos captured by these platforms usually have a high resolution with rich appearance, but with a fixed and limited spatial coverage (\textit{i.e.}, the field of view). In contrast, cameras mounted on satellites can provide spatial-temporal surveillance over a large-scale area (as shown in Fig.~\ref{thumbnail}), which is suitable for the task of urban-scale traffic management\cite{vehicle2015}, ocean
monitoring \cite{motion2017}, and smart city\cite{yang2016}. Nevertheless, the progress of moving object detection and tracking in satellite videos is still far behind its counterpart in generic videos, due to the lack of well-annotated and publicly-accessible datasets.}

\qy{With the great success of data-driven deep neural networks, remarkable progress has been achieved in the area of video understanding in recent years. In particular, several dedicated neural architectures have been proposed to tackle the problems of moving object detection \cite{moving_object_detection_survey, centernet}, visual tracking \cite{hu2017object, SiamRPN++, Siamfc, SiamRPN}, and multi-object tracking \cite{wang2018semi,luo2014multiple} in generic videos captured by commonly-used cameras. Additionally, several recent works \cite{vehicle2015,lalonde2018clusternet,improved2019,error2019, traking2017,dutracking2017,Dutracking2019,guotracking2019,xuantracking2019,shaotracking2019} have also started to generalize existing detection and tracking
frameworks to large-scale satellite videos, due to the high demand for various satellite applications such as traffic condition monitoring and forest monitoring. However, their performance is far from satisfactory \qy{due to the different nature (\textit{e.g.}, low resolution, low-contrast, and complex backgrounds) of satellite videos.}
Overall, it remains a challenging problem to extend existing moving object detection and tracking frameworks to large-scale satellite videos.
}

\qy{Achieving accurate and robust moving object detection and tracking in satellite videos is a highly challenge task due to the following reasons:\textit{First}, there is a lack of high-quality and well-annotated public datasets and comprehensive benchmarks. It is non-trivial to evaluate the performance of different algorithms in a fair and comprehensive way, especially for multi-object tracking in satellite videos. \textit{Second}, satellite videos usually have a  lower spatial resolution than generic videos. The scales of objects are very small (usually less than 50 pixels), but with large quantities (even more than 100 in a single frame). This is inherently different from generic videos captured by commercial cameras, where the objects of interest usually have rich appearance information and limited numbers in a single frame. \textit{Finally}, the background in satellite videos is more complex (with all elements in a city such as rivers, buildings, dense lanes) and continuously-changing (\textit{e.g.}, illumination variations), due to the unique viewpoint and the large field of view of the platform. Consequently, these issues pose great challenges to existing methods trained on generic videos.}

\qy{In this paper, we aim at establishing a new dataset and benchmark for moving object detection and tracking in satellite videos. In particular, we first introduce a large-scale satellite video dataset. \qy{We then propose a Motion Modeling Baseline (MMB) to achieve moving object detection on this dataset based on foreground detection and background modeling.} In addition, we further build a new satellite video benchmark to fairly and extensively evaluate the performance of existing methods in several sub-tasks, including moving object detection, single-object tracking, and multi-object tracking (Sec. \ref{subsec:eval_mot}). Finally, we identify a series of key challenges that arise in the urban-scale scenarios and provide insight into the generalization of existing pipelines to our urban-scale satellite video dataset from multiple aspects.}

\qy{Our dataset, called \textbf{VI}deo \textbf{S}atellite  \textbf{O}bjects (\textbf{\nickname}), is a large-scale dataset for moving object detection and tracking in satellite videos, which consists of 47 satellite videos captured by Jilin-1 satellite platforms\footnote{http://mall.charmingglobe.com/videoIndex.html}. As illustrated in Fig.~\ref{thumbnail}, each image has
a resolution of 12,000$\times$5,000 
and contains a great number of objects with different scales. Four common types of vechicles, including \textit{plane}, \textit{car}, \textit{ship}, and \textit{train}, are manually-labeled. A total of 1,646,038 instances are labeled by axis-aligned bounding boxes.}

\qy{In summary, the contributions of this paper can be summarized as follows}:
\begin{itemize}
\item \qy{To the best of our knowledge, the \nickname{} dataset is the \textit{first} well-annotated satellite video dataset for the tasks of moving object detection and tracking. This dataset can be used for moving object detection, single-object tracking, and multi-object tracking.}
\item \qy{We also propose \qy{\nicknamemethod{}} for tiny and moving object detection in satellite videos. Experiments show that the proposed method can effectively improve the detection rate while reducing false alarms.}
\item \qy{Extensive experiments have been conducted on the \nickname{} dataset to benchmark existing detectors and trackers. Several insightful conclusions are also drawn from the experimental results.}
\end{itemize}

\qy{The rest of this paper is organized as follows. We first give a brief introduction of related work in Section \ref{related work}. Then, we present our \nickname{} in Section \ref{dataset}, including the data collection, annotation, statistics, and properties of the dataset. Further, we introduce a dedicated \nicknamemethod{} for moving object detection based on accumulative multi-frame differencing and robust matrix completion in Section \ref{proposed method}. Additionally, we build a new satellite video benchmark for the evaluation of several sub-tasks, including moving object detection in Sec. \ref{sec:eval_detection}, single object tracking in Sec. \ref{subsec:sot-evaluation}, and multi-object tracking in Sec. \ref{subsec:eval_mot}. Finally, conclusions and suggestions for future research are given in Section \ref{conclusion}.}

\section{Related Work}\label{related work}

In this section, we briefly review the recent literature from two aspects: (1) Representative works in satellite-based object detection and tracking; (2) Existing datasets for object detection and tracking.

\begin{table*}[t]
\caption{Overview of existing satellite datasets and benchmarks for vechile detection.} \label{overview_dataset}
\centering
\resizebox{0.9\textwidth}{!}{%
\begin{tabular}{rccccccc}
\toprule[1.0pt]
 & Year & \#Images & \#Instances & \#Classes & Image width & Annotation & Sequential\\
\midrule[0.75pt]
TAS\cite{TAS}& 2008 & 30 & 1,319 & 1 & 792 & Horizontal BB & No\\
SZTAKI-INRIA\cite{INRIA} & 2012 & 9 & 665 & 1 & \textasciitilde 800 & Oriented BB & No\\
NWPU VHR-10\cite{VHR} & 2014 & 800 & 3,651 & 10 & \textasciitilde 1,000 & Horizontal BB & No\\
DLR 3K Vehicle\cite{DLR} & 2015 & 20 & 14,235 & 2 & 5616 & Oriented BB & No\\
UCAS-AOD\cite{UCAS} & 2015 & 1,510 & 14,596 & 2 & \textasciitilde 1,000 & Oriented BB & No\\
VEDAI\cite{VEDAI} & 2016 & 1,210 & 3,647 & 9 & 512/1,024 & Oriented BB & No\\
COWC\cite{COWC} & 2016 & 53 & 32,716 & 1 & 2,000-19,000 & One dot & No\\
DFC16\cite{DFC16} & 2016 & - & - & - & 3,860 & - & Yes\\
HRSC2016\cite{HRSC} & 2016 & 1,061 & 2,976 & 26 & \textasciitilde 1,100 & Oriented BB & No\\
RSOD\cite{RSOD} & 2017 & 976 & 6,950 & 4 & \textasciitilde 1,000 & Horizontal BB & No\\
CARPPK\cite{CARPPK} & 2017 & 1,448 & 89,777 & 1 & 1,280 & Horizontal BB & No\\
LEVIR\cite{LEVIR} & 2018 & 22,000 & 11,028 & 3 & 800 & Horizontal BB & No\\
VisDrone\cite{VisDrone} & 2018 & 10,209 & 54,200 & 10 & 2,000 & Horizontal BB & No\\
xView\cite{xView} & 2018 & 1,413 & 1,000,000 & 60 & \textasciitilde 3,000 & Horizontal BB & No\\
ITVCD\cite{ITVCD} & 2018 & 29,088 & 173 & 1 & 5716 & Oriented BB & No\\
DOTA-v1.0\cite{DOTA} & 2018 & 2,806 & 188,282 & 15 & 800-4,000 & Oriented BB & No\\
HRRSD\cite{HRRSD}& 2019 & 21,761 & 55,747 & 13 & 152-10,569 & Horizontal BB & No\\
DIOR\cite{DIOR} & 2019 & 23,463 & 192,472 & 20 & 800 & Horizontal BB & No\\
DOTA-v1.5\cite{DOTA} & 2019 & 2,806 & 472,089 & 16 & 800-13,000 & Oriented BB & No\\
DOTA-v2.0\cite{DOTA} & 2020 & 11,067 & 1,488,666 & 18 & 800-20,000 & Oriented BB & No\\
\textbf{\textcolor{black}{\nickname{} (Ours)}} & 2021 & 17,730 & 1,646,038& 4 &  12,000 & Horizontal BB & Yes\\
\toprule[1.0pt]
\end{tabular}
}
\end{table*}

\subsection{Satellite-based Object Detection and Tracking}
\qy{Compared to generic videos collected by ground cameras, images collected by video satellites usually have a larger field of view, with much more objects and more complex backgrounds. In addition, the objects of interest (\textit{e.g.}, cars) in satellite videos have a much lower resolution, therefore inherently lack sufficient appearance and texture information. This also poses great challenges to existing methods which heavily rely on appearance features.}

\subsubsection{Moving Object Detection in Satellite Videos}

In light of the insufficient appearance information in satellite videos, existing methods \cite{ cao2016two,leitloff2010vehicle,vehicle2015,lalonde2018clusternet,GMM,MGB,ABM,Vibe,AGMM,FRMC,FPCP,Godec,Decolor,improved2019,error2019,SptialNet,onlearning} usually achieve object detection in video sequences by utilizing motion cues or modeling the background.

In particular, frame-differencing and background subtraction are the most commonly-used frameworks. Frame differencing methods identify moving objects by thresholding the difference between consecutive frames. In particular, two-frame and three-frame differencing methods have been proposed in the literature with a number of variations \cite{ghost,fd1,fd2}. Specifically, frame-differencing methods \cite{cao2016two} leverage the motion cues in videos by detecting the changes that occurred in the foreground, further separate the foreground and background. Background modeling-based methods \cite{GMM,MGB,ABM,Vibe,AGMM,vehicle2015,FRMC,FPCP,Godec,Decolor,improved2019,error2019} first formulate the background disturbance and noise into a background model and then perform differencing with an updated model. Specifically, Kopsiaftis et al. \cite{vehicle2015} compare the change between each pixel and its surrounding pixels within two frames, and then determine whether the pixel belongs to the background. However, both background modeling and frame differencing methods require consistent global illumination and rely heavily on video frame registration.

Motivated by the enormous success of deep learning methods in various vision tasks, several works \cite{lalonde2018clusternet,onlearning} have started to leverage deep learning techniques to learn spatial-temporary cues from satellite videos. In particular, Rodney et al.  \cite{lalonde2018clusternet}  propose a two-stage framework to extract both motion and appearance information in airborne videos. A spatio-temporal CNN and a dedicated FoveaNet are used to jointly estimate the region proposals and object centroids. However, due to the lack of large-scale and well-annotated public dataset, it is still very challenging to generalize these networks to satellite videos.

\subsubsection{Object Tracking in Satellite Videos}

\qy{Due to the lack of public-available and high-quality datasets, this research topic is still in its infancy. Although several works \cite{traking2017,dutracking2017,Dutracking2019,guotracking2019,xuantracking2019,shaotracking2019} have started to transfer generic visual tracking algorithms to satellite videos, their performance on the satellite videos is far from satisfactory. Moreover, existing trackers in satellite videos are limited to large objects. There is little literature to achieve single object tracking using deep learning methods in satellite videos. In addition, Ao et al. \cite{DT}, Ahmadi et al. \cite{Ahmadi2019} and Zhang et al. \cite{zhangmot} used traditional methods to perform multi-object tracking. Therefore, a public-available and high-quality dataset is highly important in improving the performance of multi-object tracking, especially in the era of deep learning.}

\qy{To this end, we develop and release our \nickname{} in this paper, which is a new dataset and benchmark for satellite video object tracking, to foster the further development of this research area.}

\subsection{Datasets for Moving Object Detection and Tracking}

\qy{High-quality and well-annotated datasets are one of the key essential to unleashing the potential of deep neural networks.  With the revolution and great success of deep neural networks, a series of datasets for object detection and tracking have been proposed recently. As one of the representative dataset, DFC16 \cite{DFC16} is mainly used for semantic scene interpretation of space videos, including spatial scene labeling, temporal activity analysis, and traffic density estimation.} An overview of existing datasets and our \nickname{} is shown in Table \ref{overview_dataset}. Detailed attributes are also listed.

\qy{Compared to existing datasets for object detection and tracking, the proposed \nickname{} dataset has a moderate number of images and instances, slightly fewer than the very recent work DOTA-v2.0 \cite{DOTA} and xView \cite{xView}. However, the sequential nature of the data acquired by video satellites makes our \nickname{} distinguished from other existing datasets. The spatial-temporal information in our dataset provides a unique opportunity to extend existing moving object detection and tracking methods to satellite videos. On the other hand, there is no public benchmark, which is of great importance to fair and comprehensive evaluation of new algorithms, in the satellite video object detection and tracking area yet. For this reason, we establish the first benchmark for in-depth performance evaluation of moving object detection, single-object tracking, and multi-object tracking in satellite videos.}

\section{The proposed Dataset}\label{dataset}

\qy{\subsection{Data Collection and Annotation}}

\subsubsection{Videos Collection}
\qy{Our \nickname{} dataset consists of 47 satellite videos captured by the Jilin-1 satellite constellation at different positions of the satellite orbit. Jilin-1 is a video satellite system launched by the Chang Guang Satellite Technology Co., Ltd.\footnote{\href{http://www.charmingglobe.com/index.aspx}{http://www.charmingglobe.com/index.aspx}/}. The satellite videos acquired by Jilin-1 are composed of a sequence of true-color images. The covered area in real scenes is up to several square kilometers. In addition, the video frame rate is 10 frames per second. A detailed comparison between this satellite and existing satellites in terms of system configuration and image quality is shown in Table \ref{tab:Comparison-of-configuration}. In addition, we also show a series of data examples in  Fig.~\ref{example_image}, the video frame covers different types of urban-scale elements, such as roads, bridges, lakes, and various moving vehicles including cars, trains, ships, and airplanes. To build a diverse and comprehensive dataset, we also take several different traffic situations (\textit{e.g.}, dense lanes and traffic jams)}
\begin{table}[thb]
\caption{\qy{Comparison of configuration and parameters of existing satellites. SSO: Sun-Synchronous Orbit.}}\label{tab:Comparison-of-configuration}
\centering
\resizebox{0.5\textwidth}{!}{%
\begin{tabular}{rccc}
\toprule[1.0pt]
Satellite  & JiLin-1\footnote{\href{https://mall.charmingglobe.com/Sampledata}{https://mall.charmingglobe.com/Sampledata}/}  & SuperView-1\footnote{\href{http://www.spacewillinfo.com/Satellite/Satellite/superview}{http://www.spacewillinfo.com/Satellite/Satellite/superview}/} & SkySat-2\footnote{\href{https://www.youtube.com/watch?v=lKNAY5ELUZY}{https://www.youtube.com/watch?v=lKNAY5ELUZY}/} \\
\midrule[0.75pt]
Launch Year & 2015 & 2013  & 2012 \\
Spatial Resolution & 0.92 m & 0.5 m & 1.1 m\\
Standard View Size & 11 km$\times$4.6 km & 60 km$\times$70 km & 2.0 km$\times$1.1 km\\
Imaging Color & True Color & Panchromatic & Panchromatic\\
Orbit Altitude & SSO 535 km & SSO 530 km & SSO 600 km\\
Descending Node Time & 10:30 am  & 10:30 am  & 14:30 am\\
Design Life &~4 Years & ~8 Years & ~4 Years \\
Satellite Weight & 95 kg & 560 kg & 83 kg\\
Data Transmission &350 Mbps & 2$\times$450 Mbps & 450 Mbps \\
Frequency & 10 Hz & - & 30 Hz\\
Duration Time & 120 s & - & 90 s\\
\toprule[1.0pt]
\end{tabular}}

\end{table}
\qy{in real-world scenarios into consideration. Therefore, the finally selected satellite videos cover a wide range of challenges, including complex background, illumination variations, and dense lanes.}

\subsubsection{Labelling Process}
\qy{Based on discussions with industry professionals, four categories that are important for real-world applications were selected and annotated in our dataset, including \textit{car}, \textit{airplane}, \textit{ship}, and \textit{train}. Similar to \cite{krishna2017visual}, we also use 2D horizontal bounding boxes to annotate these objects \qy{for object detection}. Specifically, a common description of bounding boxes is ($x$, $y$, $w$, $h$), where ($x$, $y$) is the center of the target location, $w$, $h$ are the width and height of bounding boxes, respectively \cite{DOTA}. Although bounding boxes with orientation are used in several other datasets \cite{DOTA}, we only annotate the objects with unoriented bounding boxes. That is because the vehicles in our \nickname{} dataset are usually rigid and relatively small (as illustrated in Table \ref{instance_size}), the majority of instances in our dataset have a size smaller than 50 pixels.}

\qy{To build the tracking benchmark, we also manually label the identity of all instances. Specifically, for each instance, we assign a unique color to this instance in the first frame of the video, and then manually check all subsequent frames to ensure that the bounding boxes belong to the same object across different frames have exactly the same color (\textit{i.e.}, with the same identity). Additionally, we follow \cite{OTB} to categorize each frame in the video sequence into 7 predefined visual attributes, each attribute represents a specific challenge for object tracking. These attributes are Background Clutter (BC), Color Change (CC), Low Resolution (LR), Out-of-View (OV), Occlusion (OC), Similar Object (SOB), and Motion Blur (MB). Note, OC can further be divided into partial occlusion (POC) and full occlusion (FOC). }

\qy{In practice, we used a labeling tool\footnote{\href{https://github.com/tzutalin/labelImg}{https://github.com/tzutalin/labelImg}/} since it is convenient and can annotate multiple categories and generate XML files directly. We also provided timely feedback to the annotators and cross-checked bounding box annotations to improve the quality and consistency of annotated labels.}

\qy{\subsection{Dataset Statistics}}
\subsubsection{Image size}
\qy{Satellite video frames usually have a significantly larger spatial size than the images captured by commodity cameras. For example, the original size of images in our dataset is about 12,000$\times$5,000, while the size of most images in regular datasets  (\textit{e.g.}, NWPU VHR-10\cite{VHR}) are less than 1,000$\times$1,000. A comparison between our dataset and other object detection datasets is shown in Table \ref{comparison_data}. Considering that existing methods mainly work on images with a normal scale, we first split the satellite video frames into several similar patches. Each patch has a resolution of 1,000$\times$1,000, and is annotated and pre-processed individually.}

\begin{figure}[t]
\centering
\includegraphics[width=8.5cm]{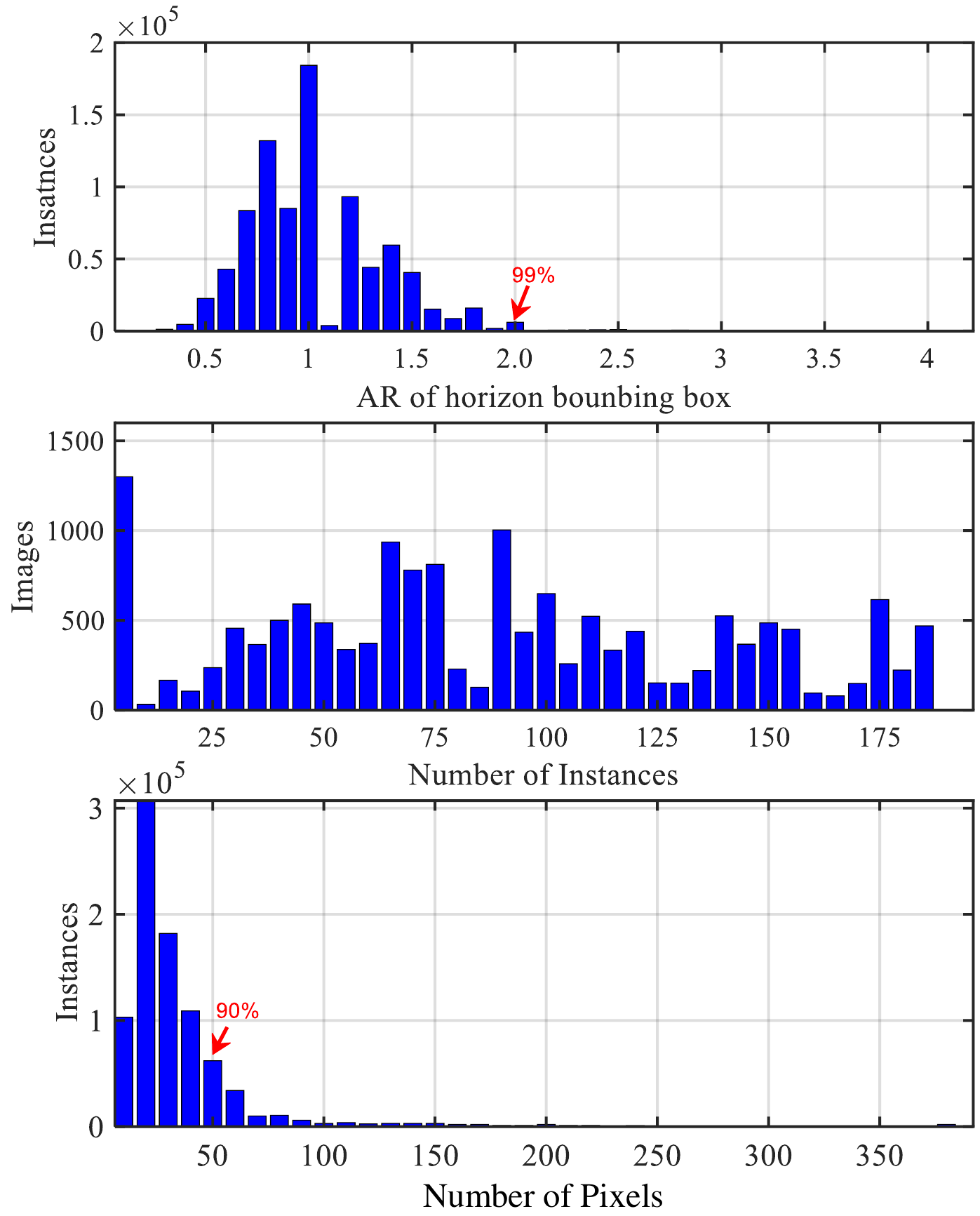}
\caption{\qy{Instance-level statistics of our \nickname{} dataset. The top row: The aspect ratio distribution of horizontal bounding boxes in our dataset. The middle row: the distribution of instance numbers of video frames in our \nickname{} dataset. The bottom row: The distribution of pixel numbers of instances in our \nickname{} dataset. }}\label{figure2}
\end{figure}

\subsubsection{Instance size}
\qy{To further compare the instance-level statistics of our dataset and other existing datasets, we follow \cite{aslani2013optical} to use the height of the horizontal border (pixel size) as the main metrics to describe the instance size. Then, we follow \cite{DOTA} to group all instances in the dataset into three subsets based on the height of the horizontal bounding box. In particular, we divide the objects into three categories according to their heights of bounding boxes: \textit{small} (10 to 50), \textit{middle} (50 to 300), and \textit{large} (above 300). Table \ref{instance_size} illustrates the instance distribution in different datasets. It is clear that the PASCAL VOC dataset \cite{everingham2010pascal} and the NWPU VHR-10 \cite{VHR} dataset are dominated by middle-size instances. In contrast, 90\% of the instances in our dataset are small objects (less than 50 pixels in scale, usually 4$\times$4 in size)}, which poses great challenges to existing methods.

\subsubsection{Instance density}

\qy{Due to the different fields of view, the number of instances per frame in satellite videos and generic videos varies greatly. For example, the average number of instances in the MSCOCO dataset is 7.7, while the number of instances in our \nickname{} dataset is up to 200. It is clear from Fig.~\ref{figure2} that the majority of images in our dataset contain a relatively large number of instances. Fig.~\ref{example_image} shows examples of densely packed instances. Detecting and tracking targets in these cases pose a great challenge for existing methods.}

\subsubsection{Instance Aspect Ratio}
\begin{figure*}[thb]
\centering
\includegraphics[width=\textwidth]{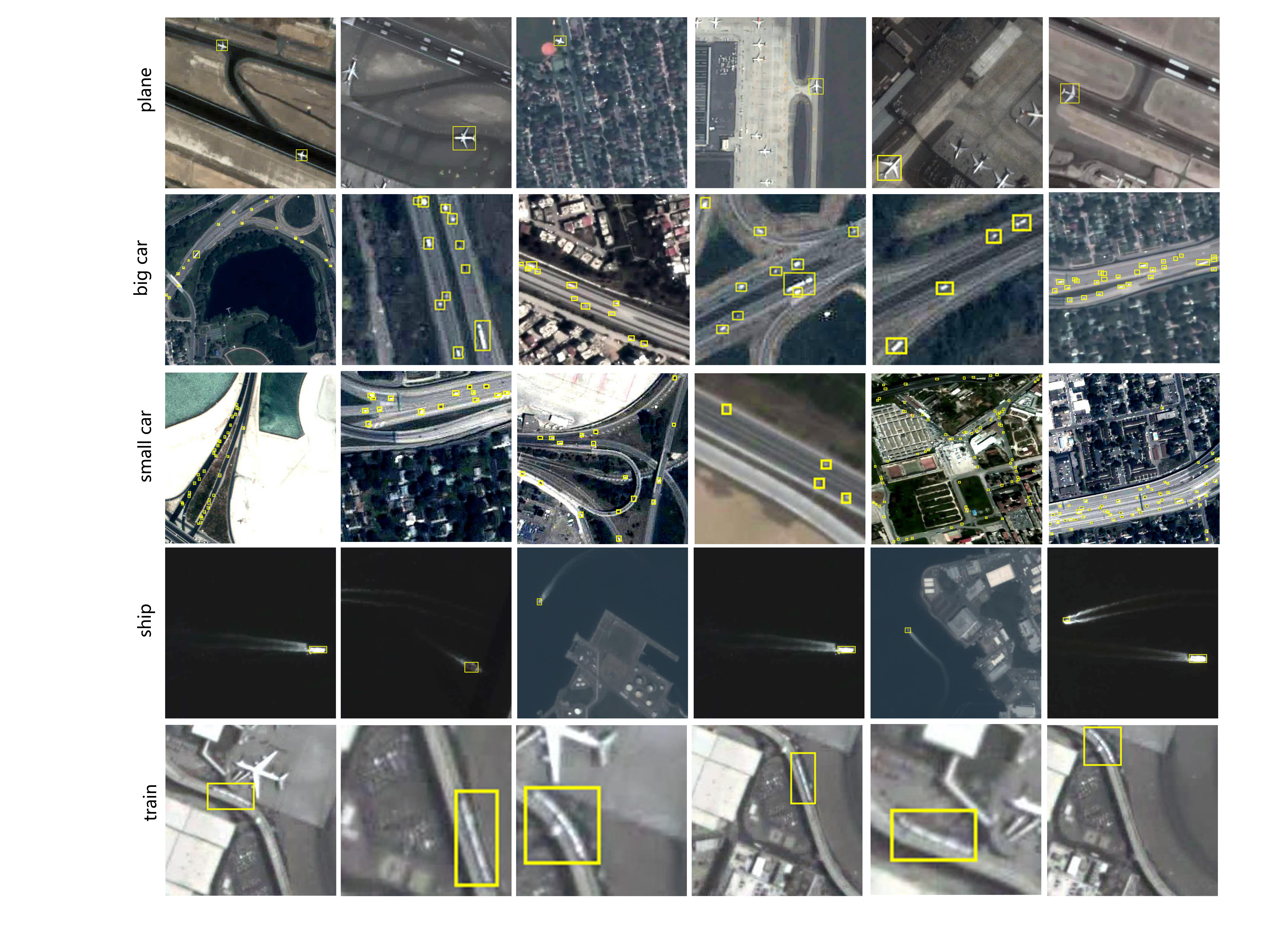}
\caption{Example images and their bounding box annotations in our \nickname{}  dataset.}\label{example_image}
\end{figure*}

\qy{For anchor-based object detection and tracking methods such as SSD \cite{ssd}, aspect ratio (AR) is an essential factor to be considered. Here, we calculate the horizontal rectangle bounding box of AR for all instances in our dataset. Fig.~\ref{example_image} illustrates the distribution of aspect ratios of all instances.}

\subsubsection{Dataset Split}


\qy{Our dataset can be used for a variety of tasks, including moving target detection, single-object tracking (SOT), and multiple object tracking (MOT). The dataset splitting is different based on the nature of each task. For the task of moving object detection, we divide the dataset into a training set (13,470 images), a validation set (535 images), and a test set (3,725 images). There are 1,646,038 bounding boxes in total. For the SOT task, the dataset provides 3159 tracklets with 1.12 million frames. Videos 1 to 20 are used to form the training set, videos 21 to 23 are used to form the validation set, while videos 24 to 27 are used to form the test set. For the task of MOT, we collected 3711 tracklets with 47 sequences, the training set is formed by videos 1 to 27 (with 3159 tracklets), the validation set is formed by videos 28 to 40 (with 440 tracklets), and the test is formed by videos 41 to 47 (with 658 tracklets).}

\begin{table}
\centering
\footnotesize
\caption{\qy{Comparison between our dataset and other existing vehicles detection datasets composed of aerial images and natural images.}}\label{comparison_data}
\resizebox{0.5\textwidth}{!}{%
\begin{tabular}{rcccc}
\toprule[1.0pt]
Dataset & Category  & Images & Image width\\
\toprule[1.0pt]
CLIF\cite{CLIF2006} & 1  & 24,564 &4,016 \\
WPAFB\cite{lalonde2018clusternet} & 1  & 1,025 &4,260 \\
NWPU VHR-10\cite{VHR} & 10  & 800 & \textasciitilde 1,000 \\
VEDAI \cite{VEDAI} & 3  & 1,268 & 512/1,024\\
3K Vehicle Detection \cite{DLR}  & 2  & 20 & 5,616\\
BIT-Vehicle \cite{BITvehicle}& 6 & 9,850&\textasciitilde 1,600/1,920\\
\textbf{\nickname{} (Ours)} & 4  & \qy{17,730} & 12,000 \\
\toprule[1.0pt]
\end{tabular}
}
\end{table}

\begin{table}
\centering
\caption{Comparison of instance size distribution of different datasets.}\label{instance_size}
\resizebox{0.5\textwidth}{!}{%
\begin{tabular}{rccc}
\toprule[1.0pt]
Dataset & 10-50 pixels & 50-300 pixels & over 300 pixels\\
\midrule[0.75pt]
PASCAL VOC (07++12)\cite{everingham2010pascal} & 14\% & 61\% & 25\%   \\
MSCOCO (2014trainval) \cite{coco}& 43\% & 49\% & 8\%   \\
NWPU VHR-10\cite{VHR} & 15\% & 83\% & 2\%   \\
DOTA\cite{DOTA} & 57\% & 41\% & 2\%   \\
\textbf{\nickname{}  (Ours)} & 90\% & 9\% & 1\%   \\
\toprule[1.0pt]
\end{tabular}
}
\end{table}

\subsection{Dataset Challenges}
\qy{Due to the unique properties of our dataset, detecting and tracking objects in our dataset has several distinctive challenges: 1) The low spatial resolution of objects of interest. We can see from Fig.~\ref{figure2} that, nearly 90\% of instances have a small number of pixels (<50). This poses a great challenge to methods which rely on appearance information. 2) The large number of instances per frame. Different from existing generic video datasets, where the number of instances in a frame is usually less than 10, our dataset has a large number of similar instances with partial occlusion, motion blur, and illumination variations.}


\section{The Proposed Method \& Baselines}\label{proposed method}
\subsection{Overview}
\qy{In this section, we will introduce a baseline method called \nicknamemethod{} (Motion Modeling Baseline) for satellite video object detection. Considering the extremely small size and limited appearance information of objects in the video satellite dataset, we first identify the key problem is: \textit{how to effectively leverage the spatial-temporal information for moving object detection in satellite videos, while suppressing false alarms in background.} Further, we propose a method based on accumulative multi-frame differencing (AMFD) and low-rank matrix completion (LRMC). As shown in Fig.~\ref{architecture}, \qy{we first propose an AMFD module to extract candidate slow-moving pixels and interest region proposals. Then, a LRMC module is introduced to detect complete objects in a computationally efficient way, which is important for large-scale satellite videos.}
Finally, a motion trajectory-based false alarm filter is proposed to reduce false alarms based on the aggregated trajectory in the time domain, since real moving objects are more likely to have a continuous trajectory.}

\subsection{Accumulative Multi-Frame Differencing}
\qy{To extract motion information (\textit{i.e.}, moving objects) from satellite videos, an intuitive and straightforward way is to calculate the difference between successive frames. Ideally, most of the unchanged background information can be eliminated. However, due to background noise and illumination variations, difference images usually contain a number of false alarms. To cope with this problem, several approaches such as symmetrical differencing \cite{qiu2014moving} are further proposed. However, these methods are unlikely to perform well on slow-moving objects with insignificant difference in appearance.}

Motivated by the use of the three-frame differencing method for moving target detection in \cite{f,fd2}, we propose a module called accumulative multi-frame differencing (AMFD) to effectively detect \qy{slow-moving} objects.
\qy{In this paper, we propose a module called accumulative multi-frame differencing (AMFD) to effectively detect \qy{slow-moving} objects. This module is based on the following motivations: (1) Due to the slow relative motion of objects across a pair of frames, it is usually difficult to detect slow-moving objects and remove the ghost generated by the edge profile of the object. (2) The object of interest (\textit{e.g.}, car, airplane) in satellite images usually has a regular geometrical pattern, texture shape, and motion models, while noise and outliers usually appear without a meaningful geometrical shape and motion pattern.}

\begin{figure*}
\centering
\includegraphics[width=16cm]{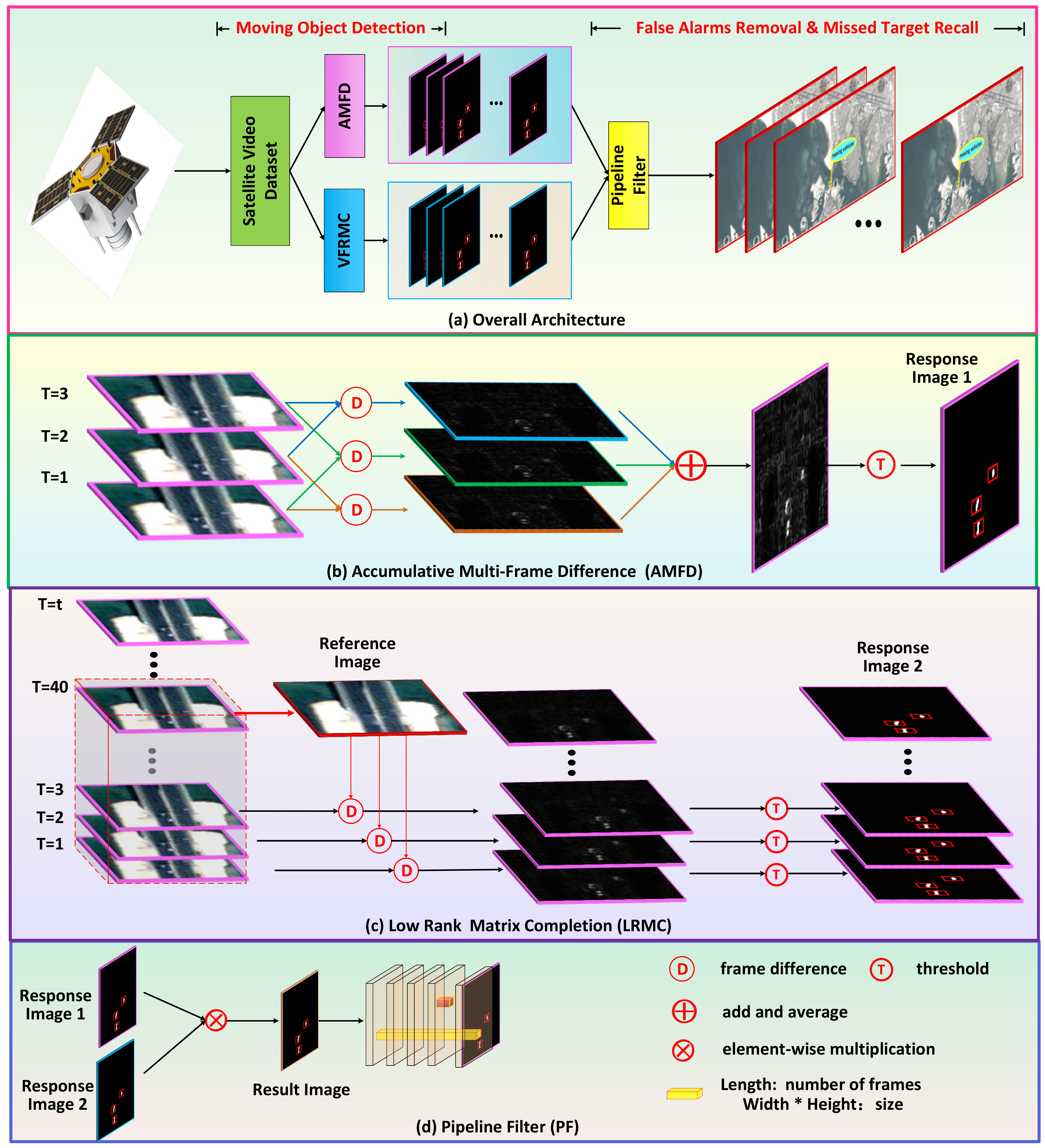}
\caption{The pipeline of the proposed \nicknamemethod{} algorithm for satellite video object detection.} \label{architecture}
\end{figure*}

\qy{Based on these motivations, we first define sub-groups using every three neighboring frames (as shown in Fig.~\ref{architecture}(b)), and then calculate the difference between frames $I_t$ and $I_{t-1}$, $I_t$ and $I_{t+1}$, as well as the difference between frames $I_{t+1}$ and $I_{t-1}$:}
\begin{equation}
D_{t1} = |I_t - I_{t-1}|, \label{eq1}
\end{equation}
\begin{equation}
D_{t2} = |I_{t+1} - I_{t-1}|, \label{eq2}
\end{equation}
\begin{equation}
D_{t3} = |I_{t+1} - I_{t}|. \label{eq3}
\end{equation}
\qy{where $D_{t1}$, $D_{t2}$ and $D_{t3}$ represent difference masks. Consequently, we can obtain a detection mask using all these detection frames.}

\qy{The detailed pipeline of our AFMD module is shown in Algorithm \ref{AMFD}. Instead of taking the intersection of two different masks{, which has been widely used in} \cite{qiu2014moving}, we further accumulate these difference masks to suppress outliers and noise, and to highlight moving objects. Specifically, three differencing masks are added {and normalized, that is:}}

\begin{equation}
I_{d} = \frac{D_{t1} + D_{t2} + D_{t3}} {3}. \label{eq0}
\end{equation}

\qy{Once the accumulative response image $I_{d}$ is obtained, we then perform binarization on this image. The pixels with values larger than threshold $T$ are kept, while others are set to zero. Due to the variation of distribution and characteristics of different response images, we determine the threshold $T$ based on the response images, rather than using a fixed threshold:
}

\qy{
\begin{equation}
I_{d}(x,y) =
\begin{cases}
255  \qquad  I_{d}(x,y) \geq T\\
0    \qquad \ \  I_{d}(x,y) < T
\end{cases} \label{eq4}
\end{equation}}

\begin{equation}
T =  \mu + k \times \sigma \label{eq5}
\end{equation}
\qy{where $\mu$ and $\sigma$ are the mean and standard deviation of the differencing image $I_{d}$, respectively. $k$ is a hyperparameter, which is empirically set to 4.}

\begin{algorithm}
\caption{AMFD based Tiny Object Detection}\label{AMFD}
\KwIn{A satellite video sequence.}
\KwOut{\qy{Candidate moving pixels}.}
\For{$i$ = 1 to $m$}
{
  \For{$j$ = $i$ to $i$+2}
 {
    calculate the differencing images $D_{t1}$, $D_{t2}$ and $D_{t3}$ according to Eqs.\ref{eq1},  \ref{eq2} and \ref{eq3}\;
    calculate the accumulative response image $I_{d}$ according to Eq.\ref{eq0}\;
    calculate a threshold $T$ to extract targets according to Eq. \ref{eq5}\;
    converted the accumulative response image to a binary image according to Eq. \ref{eq4}\;
    perform morphological operations on binary images\;
    remove false alarms according to Eq. \ref{eq6}.
 }
}
Obtain the interest region proposal and candidate moving pixels.
\end{algorithm}

\qy{\qy{In addition, due to the local misalignment and intensity changes of stationary background objects, pixel shifts of stationary background objects are likely to be considered as moving objects.} To further reduce false alarms, we incorporate prior knowledge (such as color \cite{b,c}, shape \cite{b,d},  eccentricity \cite{ghost,b}, or standard constraint (e.g., size, area) \cite{c,e,f,ghost}) of the object of interest to refine these results. In particular, size \cite{c,e,f,ghost} and aspect ratio of the objects are utilized to distinguish real targets from false ones.} In particular, the size and aspect ratio of the object are utilized to distinguish real targets from false ones. Specifically, the connected areas satisfying the following equations are considered as targets:


\begin{equation}
\begin{cases}
5\  \leq \ area \ \leq \ 80\\
1.0\ \leq\  aspect\ ratio \ \leq \ 6.0
\end{cases} \label{eq6}
\end{equation}
where the hyperparameters are empirically determined based on the statistics of the dataset.

\subsection{LRMC for Object Detection} 
\qy{The region of interest (ROI) can be calculated by AMFD. However, it is likely to detect incomplete objects and produce a large number of false targets, as mentioned in \cite{qiu2014moving}. A background subtraction method can effectively detect complete objects, this is complement with the frame differencing-based methods. Thus, another branch of our framework uses the background subtraction strategy, which aims at modeling the relative stationary background (usually cover the majority of a frame) from its input video sequence. Here, we follow the computationally efficient method \cite{FRMC} to model video frames as low-rank matrices with perturbations. Therefore, the problem of moving object detection can be transformed into recovering the low-dimensional subspace spanning the background. Specifically, the problem can be formulated as:}

\begin{equation}
{\rm min }\ rank(\bm{B}) \ {\rm s.t.}\ \bm{V} = \bm{B}+\bm{F} \label{eq7}
\end{equation}

\begin{algorithm}
\label{alg2}
\caption{LRMC based Tiny Target Detection}
\KwIn{Satellite videos.}
\KwOut{Candidate moving pixels.}
\For{$i$ = 1 to $m$}
{
    calculate the number of observation matrices according to Eq. \ref{eq8}\;
    estimate the current background model based on LRMC and obtain the foreground \;
    convert the current image into a binary image\;
    perform morphological operations to produce binary images\;
    output the candidate motion pixels.\
}
Obtain the preliminary areas with moving targets.
\end{algorithm}

\noindent \qy{ where \bm{$V$} is the observation matrix, each column of this matrix is composed of a specific video frame. \bm{$B$} is the low-rank matrix representing background. \bm{$F$} is the foreground perturbation, usually a sparse matrix. Note that, instead of using the whole video sequence to model the background \cite{FRMC}, we use sub-groups of frames to balance the number and variance of the observation matrix. In addition, a large number of observation matrices usually lead to an increase in computation without performance gain, since the difference in consecutive backgrounds is relatively small. On the other hand, a limited number of observation matrices are insufficient to model changeable backgrounds. Therefore, we determine the number of observation matrices as follows:}

\qy{\begin{equation}
N = \frac {M}{L\times f} \label{eq8}
\end{equation}
}
where $N$ is the number of observation matrices, $M$ is the number of frames in a satellite video, $f$ represents the frame frequency, and $L$ denotes the number of frames required to model background respectively. Note that, $L$ is a hyper-parameter and we empirically set $L=4$ in our framework. The detailed pipeline is shown in Algorithm \ref{alg2}.


\qy{On the other hand, we can also use the region of interest (ROI) calculated by AMFD to reduce false alarms if the background is dynamically changed.}

\subsection{Motion Trajectory based False Alarm Filter}

\qy{To further reduce false alarms and improve recall scores, we further explore the spatial and temporal information by drawing the trajectory of moving objects. Ideally, the motion trajectory of a real moving object should be continuous and regular, while the trajectory of noise should be scattered and irregular. Therefore, we propose a pipeline filter (PF) based on the motion trajectory information (MTI) to reduce false alarms.}

\qy{The overall architecture of our PF module is illustrated in Fig.~\ref{architecture}(d). For each frame, the coarse candidate pixel blobs produced by AMFD and LRMC module are fed to PF for false alarm removal. Specifically, the detailed steps of our PF with MTI method are described as follows:}



\qy{1) Parameter initialization: we empirically set the length of the pipeline as 5, and the size of the pipeline as $7\times7$. Since the size of an object is usually $3\times3$, the diameter of the pipeline is slightly larger than the target.}

\qy{2) We then perform PF starting from the first frame and its successive 5 frames. We take the first image as the current frame and identify all candidate target points $P_m$ (m=1,2,3,...) in those images.}


\qy{3) Assume $n$ objects are detected in the next frame, the objects in two frames have one-to-one correspondences. For all candidate object pixels $O_m$ in the current frame, we check whether there are object pixels in their small neighborhood in the next frame. Let $S_{ab}$ denote the Euclidean distance between the $a$-th object in the first frame and the $b$-th object in the second frame. A minimum Euclidean distance is imposed to reject assignments where the overlap between the detection and the target is larger than a threshold. $C_{ab}$ denotes the association relationship between the $a$-th object in the first frame and the $b$-th object in the second frame, the association problem can be described as follows:}

\qy{
\begin{equation}
C_{ab} =
\begin{cases}
1, \ 0 < S_{ab}(x) < 7 \ \& \ 0<S_{ab}(y) <7 \\
0, \ otherwise
\end{cases}
\end{equation}
\begin{equation}
S_{ab}(x) =  \vert{a_{x}-b_{x}}\vert
\end{equation}
\begin{equation}
S_{ab}(y) =  \vert{a_{y}-b_{y}}\vert
\end{equation}}
where $x$ and $y$ indicates the coordinates of object pixels. The optimal associations of hypotheses and ground truths
can be obtained using the Hungarian algorithm. If $C_{ab}$ = 1, we increase the object occurrence $h$ with 1. We record the position of the object pixels in the frame and set it as the current position of the candidate target point. If $C_{ab}$ = 0, this frame is skipped. Then, we further check the next frame until all frames have been iterated.

\qy{4) The number of confirmed object occurrences is counted when 5 frames are processed. If the object occurrence number \qy{$h$} is larger than \qy{$H$} (\qy{$H$} is set as 3), the candidate object is finally determined as a real object. Otherwise, it is considered as a false alarm. Besides, if \qy{$h$} equal to 3 or 4, we add the object to the frame where the object is not detected, and the position of this object can be obtained according to the motion characteristics and context information.}

\qy{5) The whole procedure is repeated until all images in the sequence are checked. }

\qy{6) The locations of all object pixels are given as outputs.}

\section{Evaluation of Object Detection}\label{sec:eval_detection}
\qy{In this section, we first introduce the baseline methods and the evaluation metrics in Section \ref{sec5.1}. Then, we present qualitative and quantitative results to demonstrate the effectiveness of our \nicknamemethod{} method in Sections \ref{sec5.2} and \ref{sec5.3}.}

\subsection{Experimental Setup}\label{sec5.1}
\subsubsection{\qy{Baseline methods}}

\qy{Our method is compared to several representative baseline methods, including two foreground detection method (\textit{i.e.}, Frame Difference (FD)\cite{cao2016two} \qy{and D\&T\cite{DT}}), six background modeling-based methods (\textit{i.e.}, ABM\cite{ABM}, MGB \cite{MGB}, Gaussian Mixture Models (GMM) \cite{GMM}, ViBe\cite{Vibe}, Regularized Background Adaptation Gaussian Mixture Models (AGMM) \cite{AGMM} \qy{and DTTP\cite{Ahmadi2019}} ), four low-rank matrix decomposition methods (\textit{i.e.}, FPCP \cite{FPCP}, Fast Low Rank Approximation (GoDec) \cite{Godec}, Contiguous Outliers in the Low-Rank Representation (DECOLOR) \cite{Decolor}, Fast Robust Matrix Completion (FRMC) \cite{FRMC}), and a deep learning method (ClusterNet) \cite{lalonde2018clusternet}}.


\begin{table*}
\centering
\renewcommand\arraystretch{1.2}
\footnotesize
\caption{Recall, Precison, and F1 values achieved by different methods on 7 satellite videos of our \nickname{}  dataset. The best results are shown in \textcolor{red}{\textbf{red}} and the second best results are shown in \textcolor{blue}{blue}.}\label{recall_precision}
\setlength{\tabcolsep}{1.8mm}{
\begin{tabular}{rccccccccccccc}
\midrule[0.75pt]
\multirow{2}*{Method} &  \multicolumn {3}{c}{Video 1}  &  \multicolumn{3}{c}{Video 2}  &  \multicolumn{3}{c}{Video 3} &  \multicolumn{3}{c}{ Video 4} \\
\cmidrule(r){2-4} \cmidrule(r){5-7} \cmidrule(r){8-10} \cmidrule(r){11-13}
& Recall & Precison  & F1   & Recall & Precison  & F1 &   Recall & Precison  & F1 &   Recall & Precison  & F1 \\
\midrule[0.75pt]
FD\cite{cao2016two} & 0.58 	&0.19 	&0.29&0.79 	&0.25 	&0.38&0.80 	&0.25 	&0.39&0.69 	&0.22 	&0.33 \\
ABM \cite{ABM} &0.81 	&0.64 	&0.71&0.79 	&0.71 	&0.75&0.92	&0.60	&0.73&\textcolor{red} {0.88} 	&0.56 	&0.68\\
MGBS \cite{MGB} &0.80	&0.47 	&0.59&0.78 	&0.52 	&0.63&0.91 	&0.36 	&0.52&0.86 	&0.27 	&0.41\\
GMM \cite{GMM}&0.37 	&0.63 	&0.47&0.49 	&0.53	&0.51&0.45	&0.53 	&0.49&0.64 	&0.36 	&0.46\\
AGMM \cite{AGMM}&0.72 	&0.56 	&0.63&0.80 	&0.77 	&0.79&0.93 	&0.65 	&0.76&\textcolor{blue}{0.87} 	&0.62 	&0.72\\
VIBE \cite{Vibe}& 0.61 	&0.34 	&0.44&\textcolor{blue}{0.82} 	&0.61 	&0.70&0.68	&0.59 	&0.63&0.65 	&0.52 	&0.58\\
FPCP \cite{FPCP}& 0.39 	&0.80 	&0.53&0.62	&0.46 	&0.53&0.82	&0.27 	&0.41&0.68	&0.22 	&0.34\\
GoDec \cite{Godec}& \textcolor{red}{0.92} 	&0.51 	&0.65&0.73 	&0.81 	&0.77&\textcolor{blue}{0.93} 	&0.53	&0.68&0.72 	&0.38 	&0.50\\
DECOLOR \cite{Decolor}& 0.24 	&\textcolor{red}{0.92} 	&0.38&0.77	&\textcolor{blue}{0.88} 	&\textcolor{blue}{0.82}&0.89	&0.83 	&\textcolor{blue}{0.86}&0.44 	&\textcolor{red}{0.93} 	&0.60\\
FRMC \cite{FRMC}& 0.55 	&0.68 	&0.62&0.57	&0.21 	&0.31&0.61	&0.21 	&0.32&0.63 	&0.17 	&0.27\\
ClusterNet \cite{lalonde2018clusternet}	&0.75	&0.67	&0.71	&0.66	&0.81	&0.72	&0.90	&0.72	&0.80	&0.50	&0.70	&0.58\\
\qy{DTTP} \cite{Ahmadi2019} & 0.74&	0.66&	0.70	&0.67&	0.84&	0.74&	0.71	&0.85	&0.77&0.64&	0.86&	0.74\\
\qy{D\&T} \cite{DT}	&0.72	&\textcolor{blue}{0.91}	&\textcolor{blue}{0.80}	&0.69	&0.86	&0.77	&0.84	&\textcolor{blue}{0.84}	&0.84	&0.76	&0.85	&\textcolor{blue}{0.80}\\

\textbf{\nicknamemethod{} (Ours)} & \textcolor{blue}{0.83} 	&0.84 	&\textcolor{red}{0.84}&\textcolor{red}{0.83} 	&\textcolor{red}{0.89} 	&\textcolor{red}{0.85}&\textcolor{red}{0.94} 	&\textcolor{red}{0.88} 	&\textcolor{red}{0.91}&0.85 	&\textcolor{blue}{0.86}	&\textcolor{red}{0.86}\\
\midrule[0.75pt]
\multirow{2}*{Method} &  \multicolumn{3}{c}{ Video 5}  &  \multicolumn{3}{c}{ Video 6}  &  \multicolumn{3}{c}{ Video 7} &  \multicolumn{3}{c}{ Average} \\
\cmidrule(r){2-4} \cmidrule(r){5-7} \cmidrule(r){8-10} \cmidrule(r){11-13}
 & Recall & Precison  & F1 &  Recall & Precison  & F1 &   Recall & Precison  & F1 &   Recall & Precison  & F1\\
\midrule[0.75pt]
FD \cite{cao2016two}& 0.61 	&0.30 	&0.47&0.80 	&0.25 	&0.39&0.80 	&0.14 	&0.24 &0.72  &0.23  &0.34\\
ABM \cite{ABM}& \textcolor{blue}{0.77} 	&0.61	&0.68&\textcolor{red}{0.83} 	&0.50 	&0.62&0.03 	&0.13	&0.04&0.72  &0.53  &0.60\\
MGBS \cite{MGB}& 0.74 	&0.39 	&0.51&0.78 	&0.24 	&0.36&0.03 	&0.13 	&0.05&0.70&0.34&0.44\\
GMM \cite{GMM}& 0.57 	&0.36 	&0.44&0.56 	&0.37 	&0.45&0.16 	&0.38 	&0.22&0.46&0.45&0.43\\
AGMM \cite{AGMM}& 0.76 	&0.68	&0.72&0.79 	&0.53 	&0.63&\textcolor{blue}{0.90} 	&0.37 	&0.53&\textcolor{blue}{0.82}&0.60&0.68\\
VIBE \cite{Vibe}& 0.72 	&0.65 	&0.69&0.60 	&0.42 	&0.49&0.45	&0.44 	&0.44&0.65&0.51&0.57\\
FPCP \cite{FPCP}& 0.33 	&0.33 	&0.33&0.65 	&0.26 	&0.37&0.68	&0.18 	&0.29&0.59&0.36&0.47\\
GoDec \cite{Godec}& 0.72 	&0.74 	&0.73&\textcolor{blue}{0.81} 	&0.42 	&0.55&\textcolor{red}{0.93} 	&0.25 	&0.39&0.82&0.52&0.61\\
DECOLOR \cite{Decolor}& 0.74 	&\textcolor{red}{0.84} 	&\textcolor{blue}{0.79}&0.71 	&\textcolor{blue}{0.80} 	&\textcolor{blue}{0.75}&0.30 	&\textcolor{blue}{0.69} 	&0.42&0.58&\textcolor{blue}{0.84}&0.66\\
FRMC \cite{FRMC}& 0.54 	&0.13 	&0.21&0.47 	&0.17 	&0.25&0.37 	&0.22	&0.28&0.53&0.26&0.32\\
ClusterNet \cite{lalonde2018clusternet}	&0.76	&\textcolor{blue}{0.82}	&0.79	&0.77	&0.71	&0.74	&0.85	&0.66	&\textcolor{blue}{0.75}	&0.74	&0.73	&0.73\\
\qy{DTTP}\cite{Ahmadi2019}& 0.62	&0.77	&0.69	&0.55	&0.73	&0.63	&0.26	&0.50	&0.34	&0.60	&0.74	&0.66\\
\qy{D\&T}\cite{DT}	&0.63	&0.81	&0.71	&0.65	&0.76	&0.70	&0.83	&0.43	&0.56	&0.73	&0.78	&\textcolor{blue}{0.74}\\
\textbf{\nicknamemethod{} (Ours)} & \textcolor{red}{0.80} 	&0.81 	&\textcolor{red}{0.80}&0.78 	&\textcolor{red}{0.85} 	&\textcolor{red}{0.81}&0.83 	&\textcolor{red}{0.73} 	&\textcolor{red}{0.78}&\textcolor{red}{0.84}&\textcolor{red}{0.85}&\textcolor{red}{0.84} \\
\midrule[0.75pt]
\end{tabular}}
\end{table*}

\subsubsection{Evaluation Metrics}

\qy{To quantitatively evaluate the performance of our method, we use six evaluation metrics, including Precision, Recall, $F_1$ score, Precision-Recall (PR) curve, Average-Precision (AP), and mean Average Precision (mAP). The detailed definition are as follows:}

\begin{itemize}
\setlength{\parsep}{0pt}
\setlength{\topsep}{0pt}
\setlength{\itemsep}{0pt}
\setlength{\parsep}{0pt}
\setlength{\parskip}{0pt}

\item \qy{Precision: For object detection, it is critical to determine whether a hypothesis is a True Positive (TP) or a False Positive (FP). Additionally, missed true targets are called False Negatives (FN). The ratio of true positives to the detected targets is defined as Precision, that is:}
\begin{equation}
\mathrm{Precision} = \frac {\mathrm{TP}} {\mathrm{TP} + \mathrm{FP}}
\end{equation}

\item \qy{Recall: This metric measures the ability of a detector to capture true targets, which is equal to the ratio of TP to the number of all existing true targets, namely:}
\begin{equation}
\mathrm{Recall} = \frac {\mathrm{TP}} {\mathrm{TP} + \mathrm{FN}}
\end{equation}

\item \qy{F1-Score: F1-score is a classical criterion for binary classification between interest targets and non-targets, which is equal to the harmonic mean of Precision and Recall, \textit{i.e.},}
\begin{equation}
\mathrm{F_1} =  \frac {2 \times \mathrm{Recall} \times \mathrm{Precision}} {\mathrm{Recall} + \mathrm{Precision}}
\end{equation}

\item \qy{Precision-Recall (PR) curve: This curve shows the tradeoff between precision and recall for different thresholds. A high area under the curve represents both high recall and high precision, where high precision relates to a low false positive rate, and high recall relates to a low false negative rate. High scores for both mean that the classifier is returning accurate results (with high precision), as well as returning a majority of all positive results (with high recall).}

\item \qy{Average-Precision (AP): It is the area under the Precision-Recall curve.}

\item \qy{mAP: It is calculated as the mean value of AP. Usually, a better classifier should have a higher AP value.}

\end{itemize}

\subsubsection{The Evaluation Protocol}
\begin{figure}
\centering
\includegraphics[width=8cm]{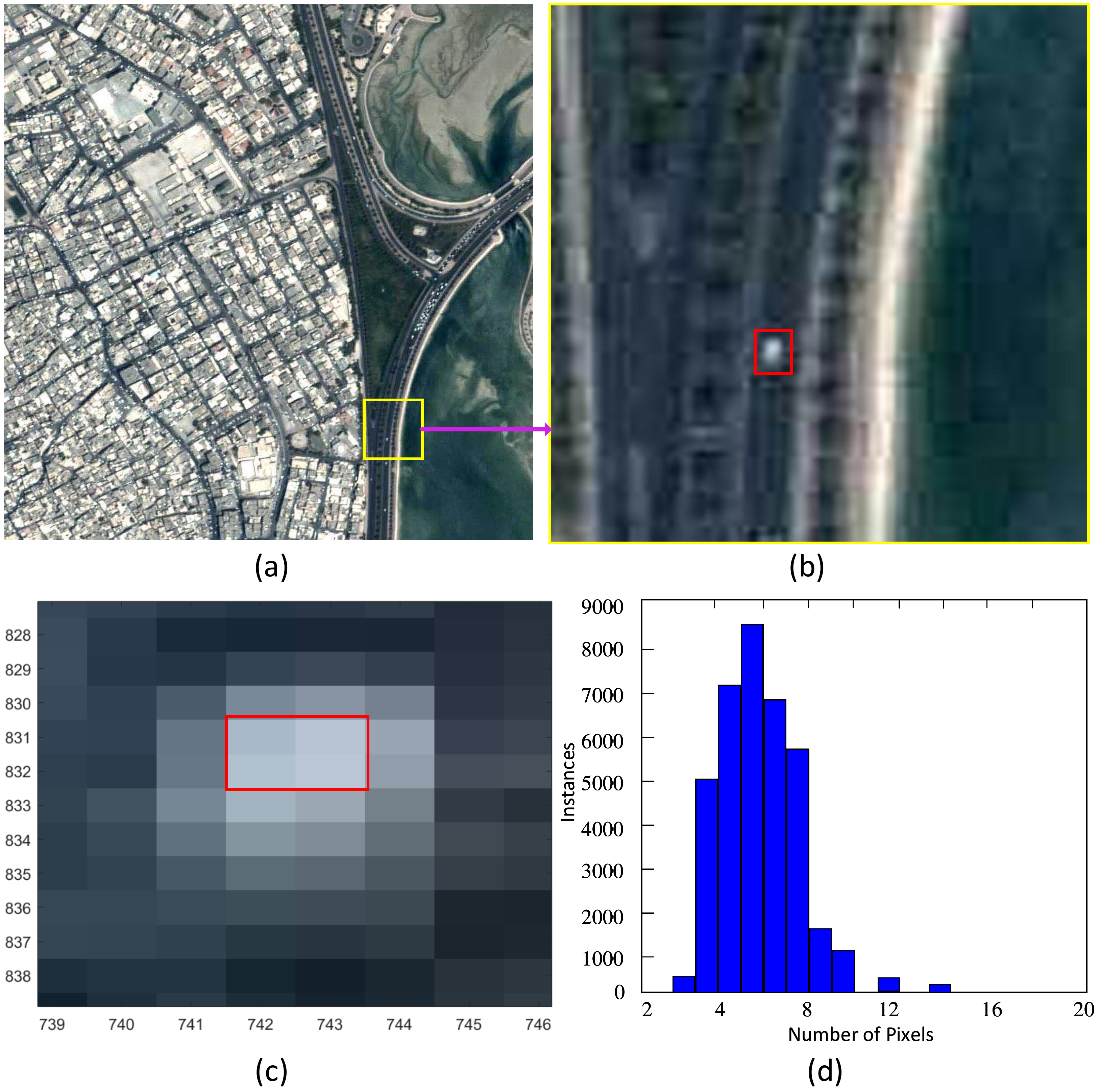}
\caption{\qy{The description of object characteristics. (a) An example frame of the proposed VISO satellite video dataset. (b) Enlarging the area within the yellow bounding box in (a) by 8 times. The red bounding box represents the ground-truth annotation of the object of interest. (c) The area within the red bounding box is further zoomed in. (d) A statistic of occupied pixel numbers for all objects in a satellite video.}} \label{target}
\end{figure}

\begin{figure}
\centering
\includegraphics[width=8cm]{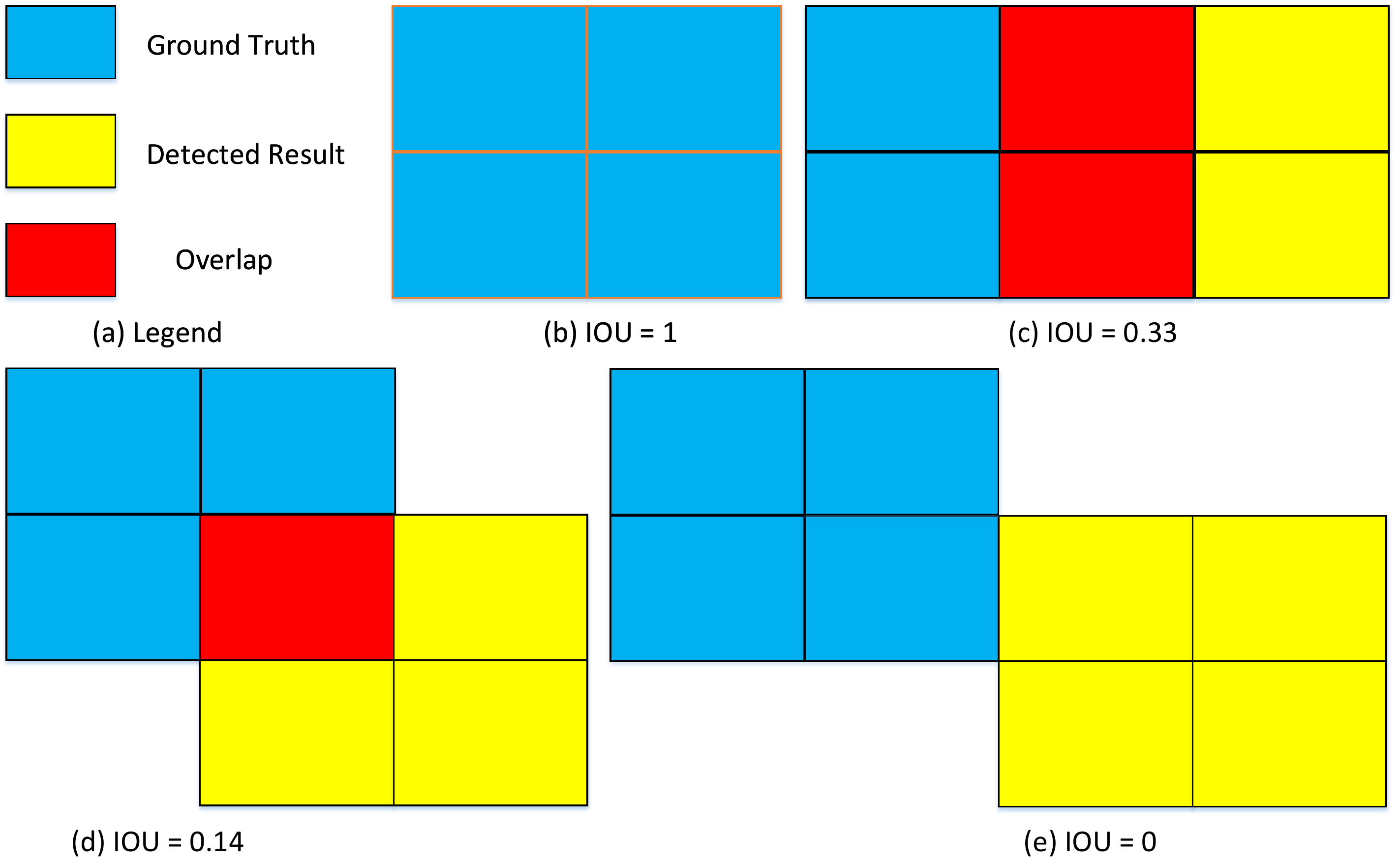}
\caption{\qy{Illustration of the IoU scores with different overlaps.}} \label{IOU}
\end{figure}

\qy{Although IoU has been widely used as an evaluation metric for generic object detection in literature \cite{bertasius2018object}, it is not particularly suitable for our case, due to the low spatial resolution of satellite remote sensing videos and extremely small objects. As shown in  Fig.~\ref{target}(d), the majority of the vehicles in our dataset only occupy 2\textasciitilde20 pixels. In this case, the conventional IoU metric is quite sensitive to the predictions. Taking Fig.~\ref{IOU} as an example, the object size is around 4 pixels, and tiny shifts of the predicted bounding box (\textit{i.e.}, 1 or 2 pixels) will cause a large fluctuation of the IoU score. Therefore, we instead consider a predicted detection as a true positive if the predicted bounding box is overlapped with the ground truth bounding box.}

\begin{table*}
\centering
\renewcommand\arraystretch{1.2}
\footnotesize
\caption{AP and mAP values achieved by different methods on 7 satellite videos  of our \nickname{}  dataset. The best results are shown in \textcolor{red}{\textbf{red}} and the second best results are shown in \textcolor{blue}{blue}.\label{AP_MAP}}
\setlength{\tabcolsep}{4.2mm}{
\begin{tabular}{rcccccccc}
\midrule[0.75pt]
Method & Video 1  & Video 2 & Video3 & Video 4 & Video 5 & Video 6 & Video 7 & mAP \\
\midrule[0.75pt]
FD \cite{cao2016two}&0.26 &  0.55&  0.51&  0.47  & 0.44 & 0.55 & 0.28  & 0.43\\
ABM \cite{ABM}& 0.59&	0.70&	0.76&	0.67&	0.65&	0.63&	0.01&0.57 \\
MGBS \cite{MGB}&0.73& 0.62& 0.58&0.41& 0.49& 0.30&0.10& 0.46\\
GMM \cite{GMM}&  0.41&0.41&0.56&0.39&0.35&0.36&0.16&0.38 \\
AGMM \cite{AGMM}&  0.64&0.71&0.75&0.71&0.61&0.60&0.47&0.64\\
VIBE \cite{Vibe}& 0.55&0.73&0.56&0.54&0.67&0.50&0.30&0.55\\
FPCP \cite{FPCP}& 0.54&0.45&0.42&0.28&0.27&0.35&0.20&0.36\\
GoDec \cite{Godec}& 0.56&0.75&0.72&0.64&0.74&0.68&0.30&0.63\\
DECOLOR \cite{Decolor}&0.42&0.73&\textcolor{blue}{0.81}&0.56&0.75&0.72&0.52&0.64\\
FRMC \cite{FRMC}& 0.51&0.22&0.23&0.20&0.11&0.14&0.13&0.22\\
ClusterNet \cite{lalonde2018clusternet}& 0.77&0.73&0.74&0.61&\textcolor{blue}{0.77}&\textcolor{blue}{0.75}&\textcolor{blue}{0.57}&0.71\\
DTTP\cite{Ahmadi2019}& 0.75&0.72&0.76&0.63&0.65&0.69&0.42&0.66\\
D\&T\cite{DT}& \textcolor{blue}{0.79}&\textcolor{blue}{0.76}&0.80&\textcolor{blue}{0.81}&0.73&0.70&0.54&\textcolor{blue}{0.73}\\
\textbf{\nicknamemethod{} (Ours)}  & \textcolor{red}{0.88} & \textcolor{red}{0.84} & \textcolor{red}{0.95} &   \textcolor{red}{0.89}  & \textcolor{red}{0.81} &   \textcolor{red}{0.77} &  \textcolor{red}{0.73} &   \textcolor{red}{0.83} \\
\midrule[0.75pt]
\end{tabular}}
\end{table*}
\begin{figure*}[t]
\centering
\includegraphics[width=18cm]{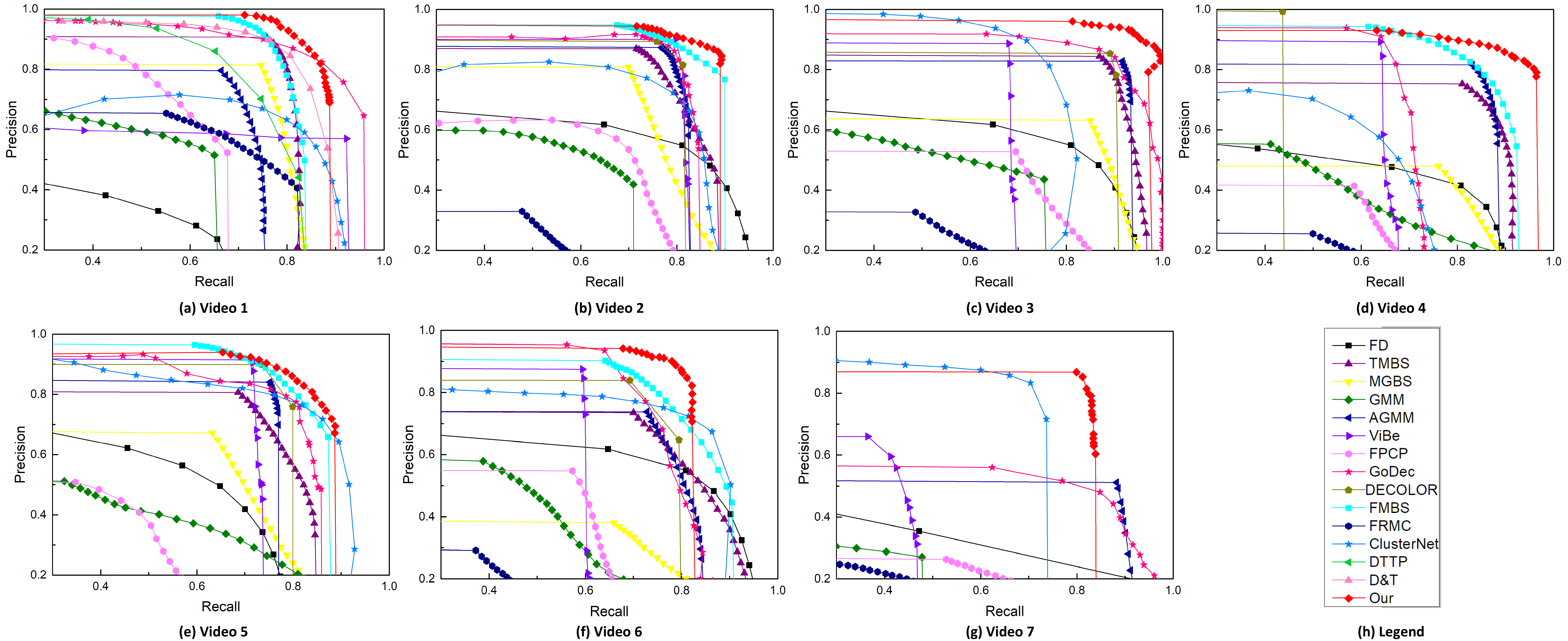}
\caption{PR curves achieved by different methods on videos 1-7 of our \nickname{}  dataset.\label{pr_curve}}
\end{figure*}
\subsection{Quantitative Results}\label{sec5.2}

\qy{Here, we evaluate the proposed method and other baselines on seven representative videos selected from our \nickname{} dataset using these evaluation metrics. Note that, these results are achieved using all frames in each video sequence. Quantitative results achieved by our method and other baseline methods are summarized in Tables \ref{recall_precision} and \ref{AP_MAP}. It can be seen that: \textit{First}, our method achieves the highest $F_1$ score on all of the 7 videos, with a large performance improvement compared to existing methods. In particular, our method outperforms the second top-performing method \qy{D\&T \cite{DT}} by \qy{10\%} in terms of average $F_1$ score. \textit{Second}, the proposed approach also achieves the highest Recall and Precision scores in videos 2 and 3, and outstanding results on other videos, leading to the top-performing results in terms of average Precision and Recall. \textit{Finally}, our method outperforms all existing methods in terms of mAP value, with a score of 0.83. In particular, our method can identify slow-moving objects from complex backgrounds with fewer false alarms, hence is more effective than temporal-based methods such as FD \cite{cao2016two}. }

{On the other hand, our method is less sensitive to the changing background compared to other vanilla background modeling methods, such as ABM \cite{ABM} and GMM \cite{GMM}.} 
\qy{We also show the PR curves achieved by our method and other baselines in Fig.~\ref{pr_curve}. It can be seen that the PR curves achieved by our method on all these 7 satellite videos are close to the upper right corner. This means our method outperforms other baseline methods in terms of both Precision and Recall. It is also noted that the gray values of objects are very close to the background in video 7. Therefore, the changing area cannot be accurately detected by several existing methods, such as FPCP \cite{FPCP} and FRMC \cite{FRMC}. As a result, the PR curve of these methods on video 7 is missing. In contrast, our approach still achieves a satisfactory performance on video 7, this further illustrates that our method is more suitable for slow-moving object detection in various complex and changing backgrounds. }

\qy{Overall, our method can achieve the best performance on all of those videos. This further illustrates the high accuracy and robustness (with less false alarms) of our method.}

\begin{figure*}[t]
\centering
\includegraphics[width=18cm]{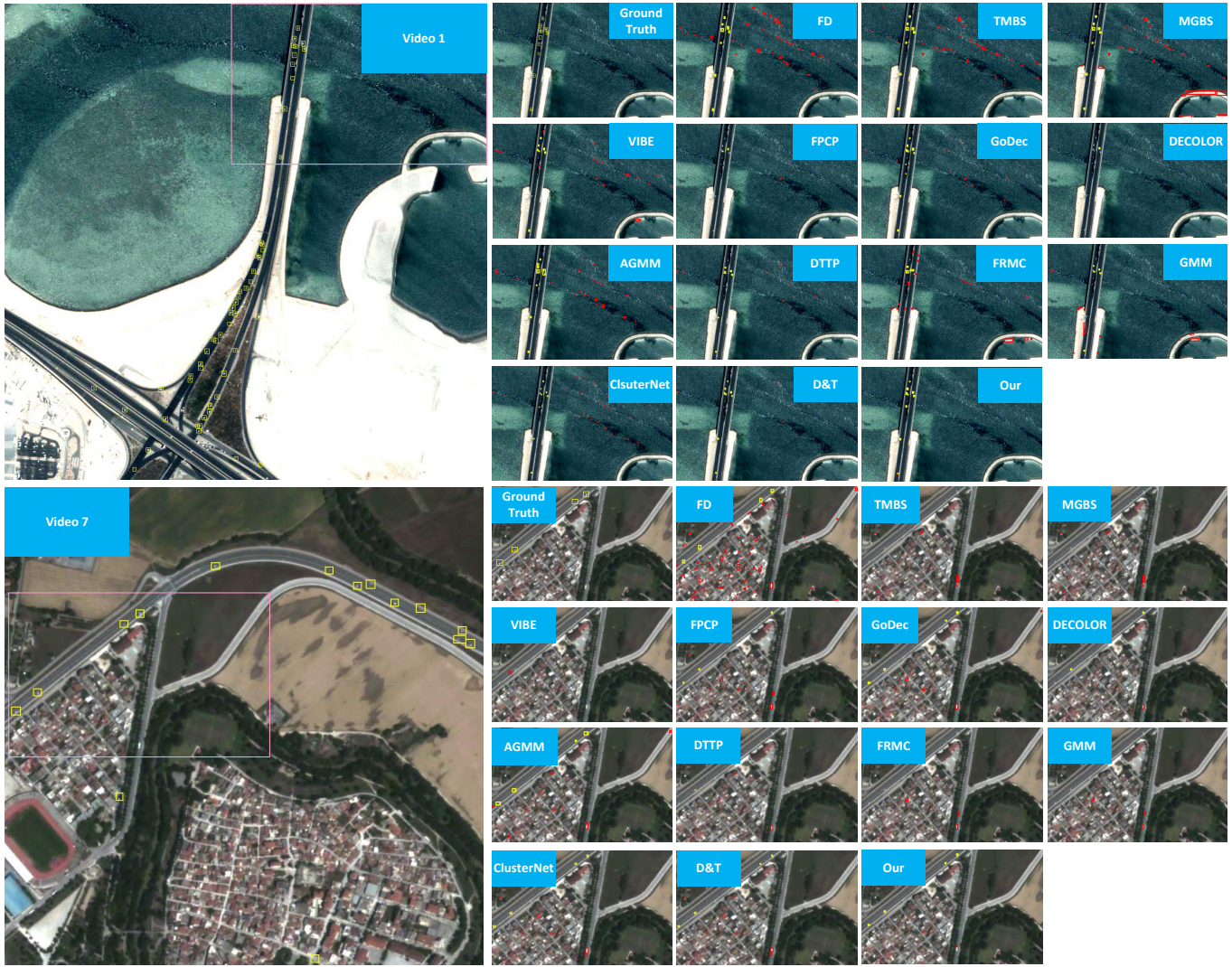}
\caption{Target detection results obtained by different methods. Ground-truth targets are shown in pink rectangles, a close-up version is shown in the each figure, positive targets are shown in yellow rectangles, false alarms are shown in red rectangles. \label{qualitative}}
\end{figure*}

\subsection{Qualitative Results}\label{sec5.3}

Qualitative results achieved by our method and other baseline methods on videos 1 and 7 are shown in Fig.~\ref{qualitative}. The ground-truth objects are shown in yellow rectangles, and positive detection results and false alarms are shown in the yellow and red rectangles, respectively. For more qualitative results, please refer to \url{https://github.com/QingyongHu/VISO}. We can see from the zoomed-in qualitative results that, the objects of interest in video 1 move relatively slowly, but the intensity value of the water surface changes significantly due to illumination variations. As a result, several baseline methods (\textit{i.e.}, FD, ABM, MGBS, GMM, AGMM, and ViBe) produce a considerable number of false alarms.  In video 7, although FPCP, GoDec, Decolor, and FRMC can decrease the false alarms to a certain extent, they also fail to detect the objects of interest due to the low contrast of objects to the background and low Signal-to-noise ratio (SNR). Different from these methods, our method is able to successfully detect slow-moving objects in both scenarios, but with much fewer false alarms. This can be attributed to the accumulative multi-frame differencing module and the effectively low-rank matrix completion module.

\qy{\subsection{Ablation Study}}
\qy{In this subsection, we compare our \nicknamemethod{} with several variants to investigate the potential benefits introduced by our network modules. As shown in Table \ref{Ablation}, the detection performance in terms of precision, recall, and F1 score is improved significantly when utilizing both the AMFD, LRMC, and PF modules, demonstrating that the combination of these modules is effective. In addition, the utilization of post-processing modules (\textit{i.e.}, PF) can also improve recall and reduce false alarms.}

\subsection{Parameter Choice}
\begin{figure}[t]
\centering
\includegraphics[width=8cm]{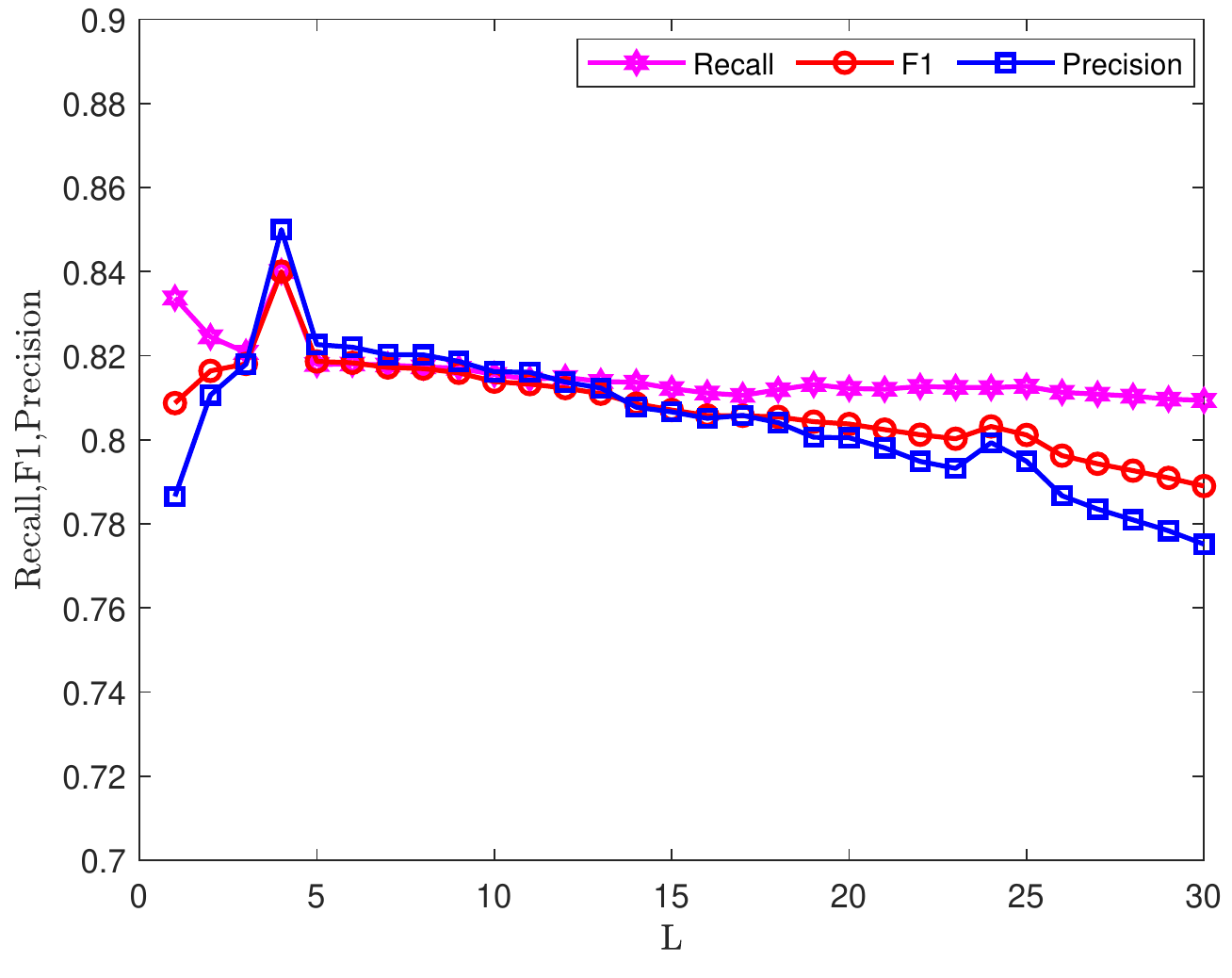}
\caption{\qy{Precision, recall, and F1 vary with the value of $L$. }}\label{L}
\end{figure}

\qy{To justify the parameter choice of $L$ in our framework, we further conduct experiments on the proposed dataset. Specifically, we show the experimental results including Precision, Recall, and F1 scores in Fig.~\ref{L} by varying the parameter $L$ from 1 to 30. We can see that the best performance is achieved when $L$ is set to 4. Therefore, we simply set $L=4$ in our framework. }

\begin{table}
\centering
\renewcommand\arraystretch{1.2}
\footnotesize
\caption{ Recall, Precision and F1 values on 7 videos achieved by different variants.\label{Ablation}}
\setlength{\tabcolsep}{4.2mm}{
\begin{tabular}{rcccc}
\midrule[0.75pt]
Module & Recall   & Precision & F1  \\
\midrule[0.75pt]
AMFD only &0.83 &  0.66&  0.73 \\
LRMC only & 0.82&	0.67&	0.74\\
Removing PF &0.84& 0.75& 0.80\\
\textbf{\nicknamemethod{} (Ours)}  & \textcolor{red}{0.85} & \textcolor{red}{0.84} & \textcolor{red}{0.85}  \\
\midrule[0.75pt]
\end{tabular}}
\end{table}

\section{Evaluations of Object Tracking}
\qy{The proposed \nickname{} dataset can  also be used to test the performance of existing tracking algorithms in satellite videos. In particular, we build the \textit{first} benchmark for the evaluation of single object tracking (SOT, See Section \ref{subsec:sot-evaluation}) and multiple-object tracking (MOT, See Section \ref{subsec:eval_mot}) algorithms in satellite videos.}


\subsection{Evaluation of Single-object Tracking Algorithms} \label{subsec:sot-evaluation}
\qy{The first question here is: \textit{Can existing single-object tracking algorithms for generic video sequences be generalized to satellite video sequences?} especially when the objects in satellite videos are far smaller than the objects in normal images. To answer this question, we carefully selected ten representative trackers and evaluated their performance on our \nickname{}  dataset. These methods can be divided into two categories: (1) Correlation Filter (CF) based-trackers: KCF \cite{KCF}, fDSST \cite{fDSST}, ECO \cite{ECO}, MCCT \cite{MCCT}, STRCF \cite{STRCF}, \qy{and CFME\cite{xuantracking2019}}. These methods are usually built upon hand-crafted feature descriptors such as HOG and color names. (2) Siamese network-based trackers: SiamFC \cite{Siamfc}, SiamRPN \cite{SiamRPN}, DaSiamRPN \cite{DaSiamRPN}, SiamRPN++ \cite{SiamRPN++}, and SiamBAN \cite{SiamBAN}. Note that, these Siamese networks are trained on generic video datasets (\textit{e.g.}, ImageNet VID \cite{ImageNetVID} ) due to the limited number of satellite video sequences.}



\subsubsection{Experimental Settings} \qy{We adapted our \nickname{} dataset for single object tracking by generating a tracklet for each individual instance in these sequences. Similar to the standard visual tracking scheme \cite{OTB}, we only provided the ground-truth bounding box of the objects at the first frame for initialization. In total, we collected 3,159 tracklets with 1.12 million frames. Compared to existing tracking datasets, our dataset contains common object tracking challenges such as illumination variation, fast motion, similar appearance, and full occlusion. Moreover, constantly low-resolution and insufficient information widely exist in our dataset due to the nature of satellite images, which is different from existing datasets.}

\subsubsection{Evaluation Metrics} \qy{Following the standard evaluation protocol of the OTB \cite{OTB} dataset, all trackers are evaluated using two metrics: Distance Precision Rate (DPR) and Overlap Success Rate (OSR). DPR is measured as the percentage of frames whose center location errors (\textit{i.e.}, Euclidean distances between centers of the predicted box $B_p$ and the ground-truth $B_g$) are smaller than a given threshold $\alpha$. Note that, we use the DPR score for the threshold $\alpha=5$ pixels since most instance objects are small and only occupied 10-50 pixels (See Fig.~\ref{figure2}). OSR is shown as the percentage of frames whose overlap ratios with the ground-truth box are larger than another given threshold of $\beta$. The overlap ratio is defined as $S=\frac{|B_p\cap{B_g}|}{|B_p\cup{B_g}|}$, where $|\cdot|$ denotes the number of pixels in a region, $\cap$ and $\cup$ denote the intersection and union of two regions, respectively.}

\begin{table}
\centering
\caption{DPR and OSR scores achieved by different methods on our \nickname{} dataset. The best results are in \textcolor{red}{\textbf{red}} and the second best results are in \textcolor{blue}{\textbf{blue}}. (HOG: Histogram of oriented gradient, CN: Color names, ConvFeat: Convolutional feature.) \label{results-SOT}}
\resizebox{0.5\textwidth}{!}{
\begin{tabular}{rcccc}
\toprule[1.0pt]
Method & Features & DPR (\%) & OSR (\%) & Speed (fps) \\
\toprule[1.0pt]
KCF \cite{KCF} & HOG & 11.5 & 5.0 & \textcolor{red}{1668.3} \\
fDSST \cite{fDSST} & HOG & 21.0 & 10.0 & 320.5\\
ECO \cite{ECO} & HOG+CN & \textcolor{red}{61.2} & \textcolor{blue}{34.5} & 48.1 \\
MCCT \cite{MCCT} & HOG+CN & \textcolor{blue}{60.5} & \textcolor{red}{34.7} & 42.8 \\
STRCF \cite{STRCF} & HOG+CN & 57.7 & 29.2 & 44.4 \\
CFME \cite{xuantracking2019} & HOG & 50.4 & 28.2 & \textcolor{blue}{422.8} \\
\midrule[0.5pt]
SiamFC \cite{Siamfc} & ConvFeat (AlexNet) & 49.1 & 26.9 & 30.1 \\
SiamRPN \cite{SiamRPN} & ConvFeat (AlexNet) & 43.7 & 17.1 & 187.4 \\
DaSiamRPN \cite{DaSiamRPN} & ConvFeat (AlexNet) & 49.1 & 19.3 & 86.4\\
SiamRPN++ \cite{SiamRPN++} & ConvFeat (ResNet50) & 47.9 & 19.9 & 45.1\\
SiamBAN \cite{SiamBAN} & ConvFeat (ResNet50) & 48.0 & 18.8 & 64.2 \\
\toprule[1.0pt]
\end{tabular}}
\end{table}

\begin{figure}[t]
\centering
\includegraphics[width=0.5\textwidth]{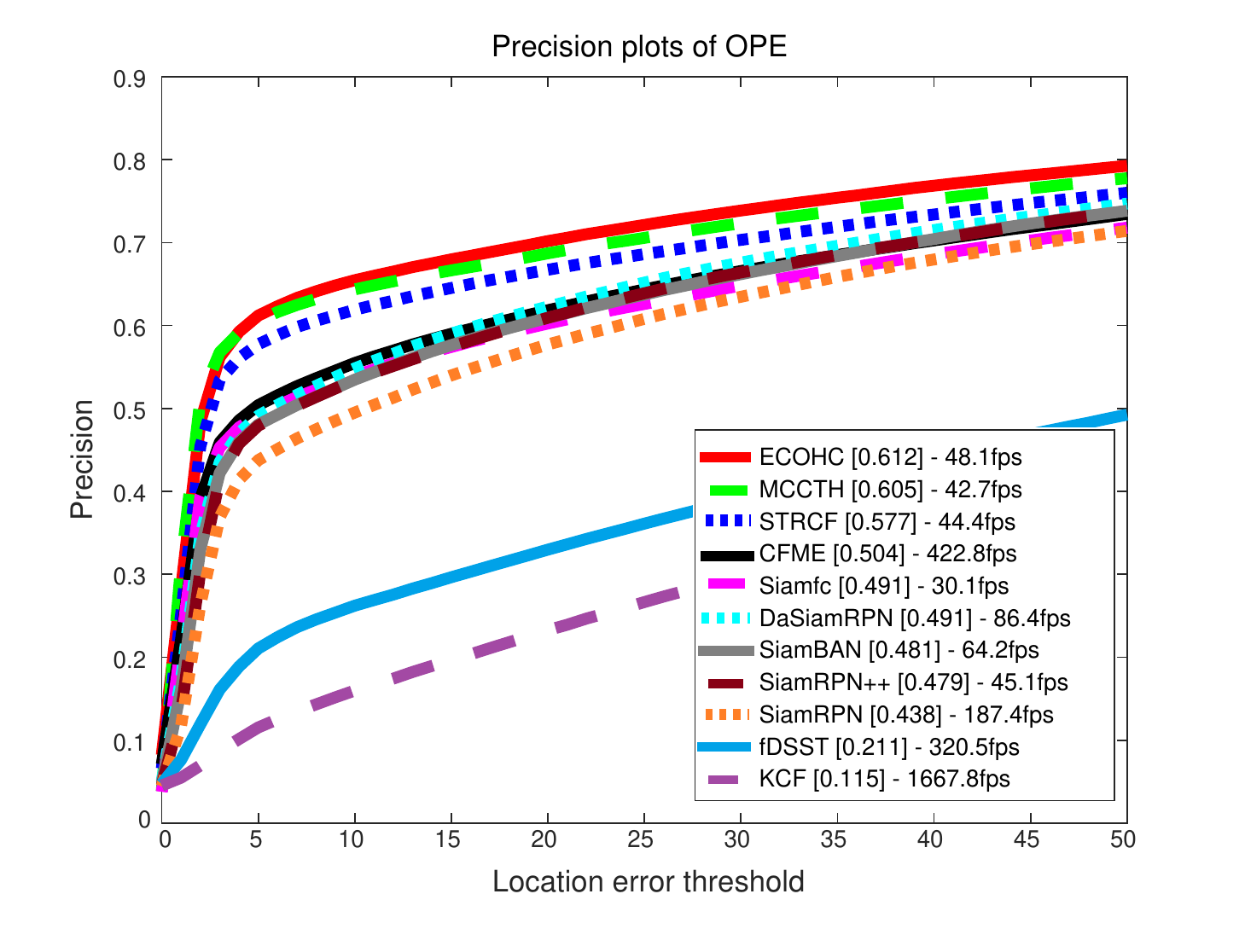}
\caption{Experimental results of moving vehicles} tracking. Precision plots over all the sequences and the legend of the precision plot is the precision score for each tracker. \label{sot_d}
\end{figure}
\begin{figure}[t]
\centering
\includegraphics[width=0.5\textwidth]{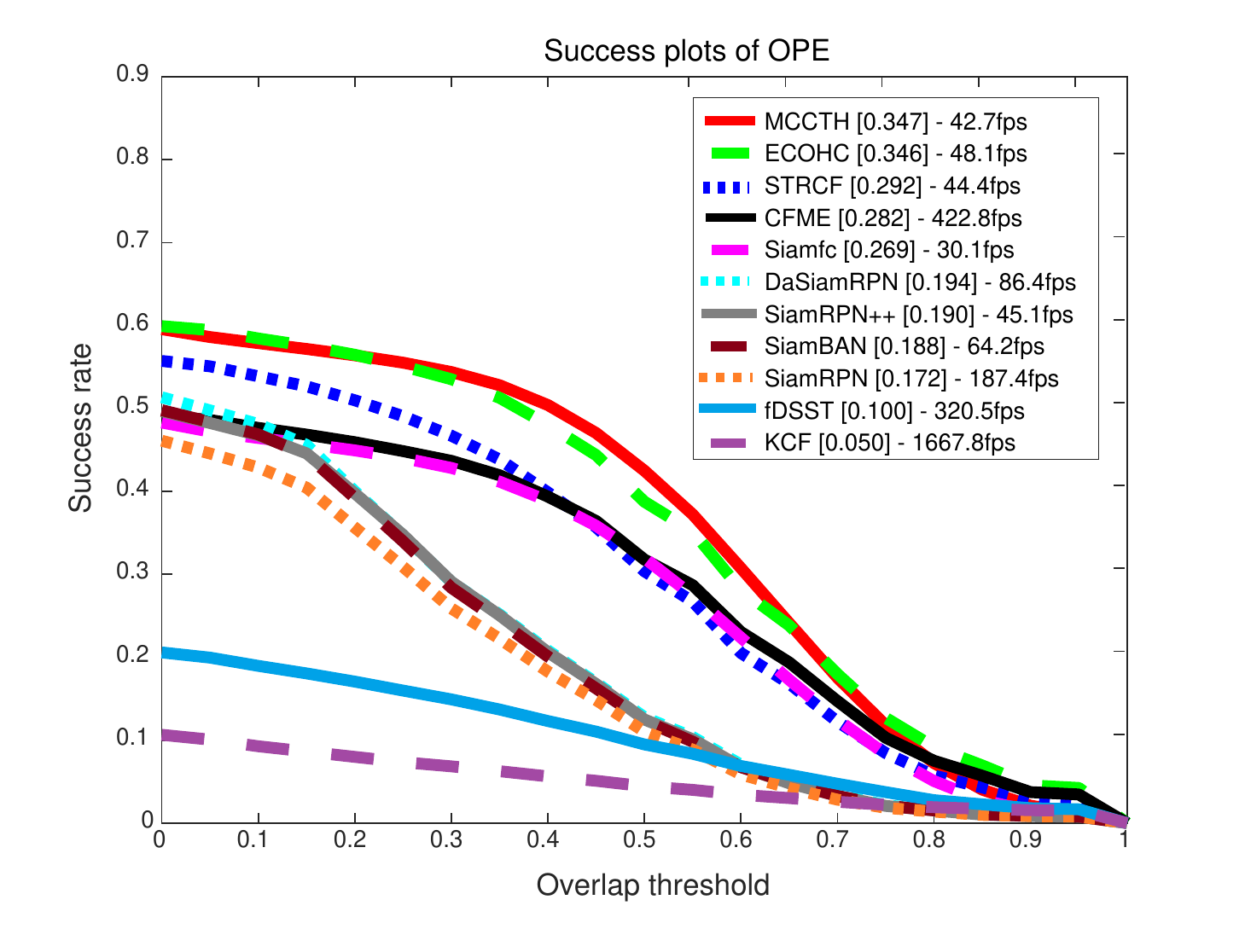}
\caption{Experimental results of moving vehicles tracking. Success plot over all the sequences and the legend of the success plot is the AUC for each tracker. \label{sot_o}}
\end{figure}

\begin{figure}[t]
\centering
\includegraphics[width=0.5\textwidth]{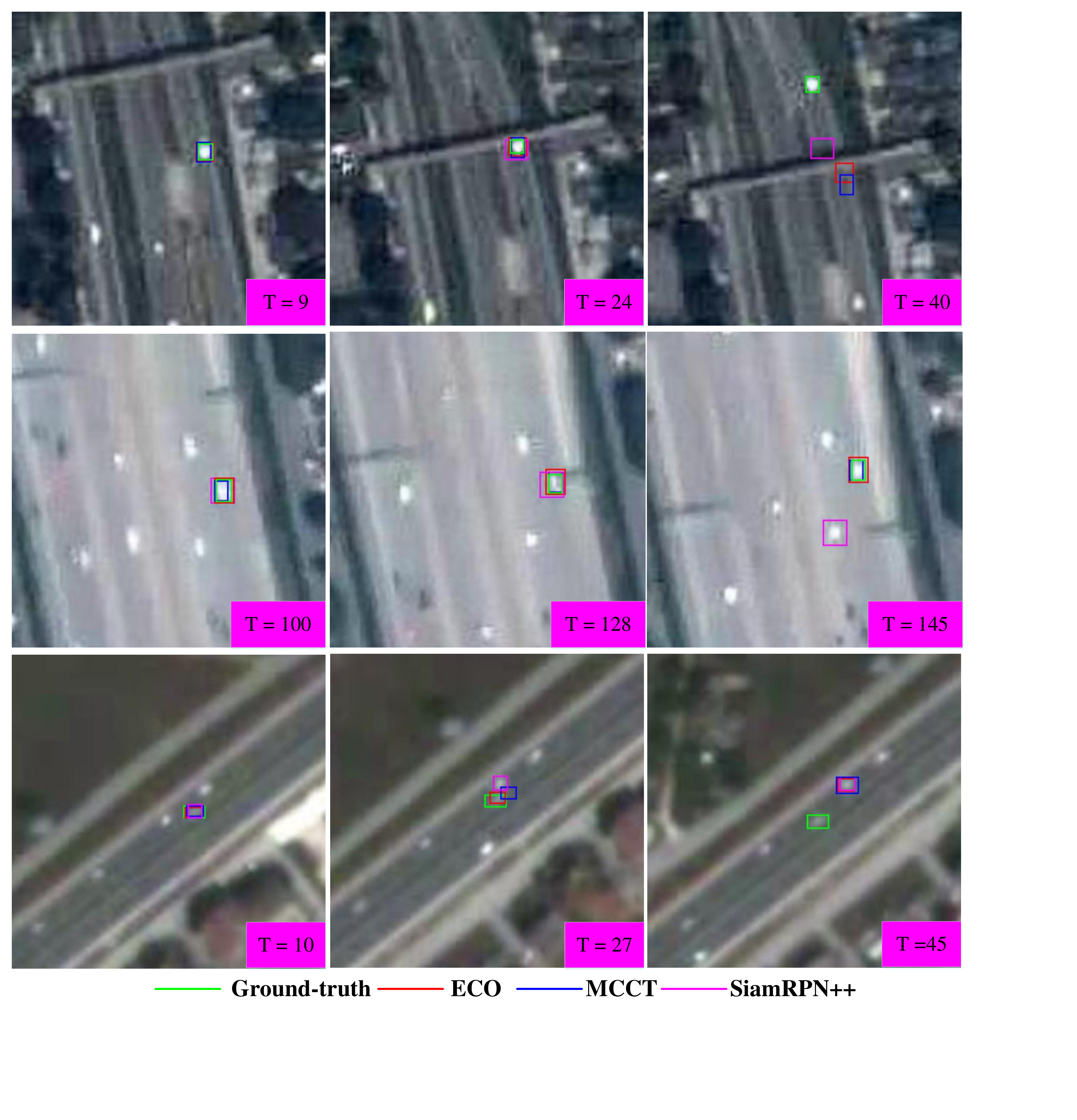}
\caption{Qualitative results achieved by ECO, MCCT and SiamRPN++. We use different colors to show the results predicted by different trackers. \label{sot_vis}}
\end{figure}

\subsubsection{Experimental Analyses}

\qy{We show the DPR, OSR, and tracking speed results achieved by these eleven trackers in Table \ref{results-SOT}, Fig.~\ref{sot_d} and Fig.~\ref{sot_o}. We can see that ECO achieves the best DPR score of 61.2\%, MCCT obtains the best OSR score of 34.7\%. CFME \cite{xuantracking2019} is a dedicated approach to achieve object tracking in remote sensing videos. However, the tracking performance on our dataset is far from satisfactory, primarily because most of the objects in our dataset are extremely small (less than 3$\times$3) and low resolution. Interestingly, we also find that the performance of SiamRPN++ and SiamBAN (which are the top-performing trackers on generic video object tracking datasets) show a significant decrease on our dataset, demonstrating that the domain gap between generic videos and satellite videos needs further exploration.}

\qy{To have an intuitive and qualitative comparison of different trackers, we visualize the tracking results achieved by several trackers (\textit{e.g.}, ECO, MCCT, and SiamRPN++) on three challenging sequences, as shown in Fig.~\ref{sot_vis}. For more qualitative results, please refer to \url{https://github.com/QingyongHu/VISO}. These sequences contain several satellite video challenges, including low resolution, similar object distractor, fast motion, and occlusion. We can observe from Fig.~\ref{sot_vis} that these trackers are prone to lose the target in scenes with occlusion. A potential solution for generic object tracking is to leverage an instance-specific detector to re-locate the target. Further, video sequence with fast object motion is difficult due to the limit in the size of the search region (which is usually proportional to the size of the target).}

\qy{For future research, we believe that it is worthwhile to further investigate the approach for effective incorporation of motion and historical trajectory information in visual trackers, as the images from satellite platforms inherently lack sufficient appearance and texture information. On the other hand, using super-resolution methods\cite{SOFVSR2020,Dong2020Remote,Dong2020Remotea} to improve image quality before object tracking is also a potentially feasible solution.}

\subsection{Multi-object Tracking Evaluation}\label{subsec:eval_mot}
\qy{Another question we would like to investigate is: \textit{Can we achieve accurate multi-object tracking in satellite videos?} To this end, we tested several representative multiple-object tracking algorithms on our \nickname{} dataset. In particular, we  selected six representative MOT algorithms for performance evaluation, \qy{including Kalman Filter\cite{KF}, CMOT \cite{TC}, SORT\cite{Sort}, FairMOT\cite{FairMOT}, DTTP\cite{Ahmadi2019}, and D\&T\cite{DT}}. Note that, the inputs (\textit{i.e.,} detection results at each frame) to these baselines are the same for fair comparison. Here, we used the detection results achieved by our \nicknamemethod{} method.}


\subsubsection{Experimental Settings}

\qy{To adapt the dataset to multi-object tracking settings, we manually associated the same objects across the video sequences to generate a series of individual tracklets. For the seven satellite videos captured by Jilin-1, we obtained 658 tracklets with 89,509 bounding boxes. Different from the single-object tracking experiments in Sec. \ref{subsec:sot-evaluation}, there are two additional and unique challenges that need to be further solved for multiple-object tracking: 1) The number of the objects in a scene usually varies over time; 2) The identity (object id) of each object needs to be fixed, otherwise, multiple-object tracking would be easily failed. To further illustrate these two subtasks: we show several representative scenarios of our dataset in Fig.~\ref{mot_vis}. It is clear that there can be dozens of objects in a single frame, and each object should have  its unique identity. In addition, the objects of interest may move out of the scene (termination) or re-enter (initialization) at any time, which makes the task much more challenging. For these baselines, we used the same parameters as their original implementations with minimal modifications. All these experiments were conducted on a PC with 3.00 GHz CPUs and 8GB RAM. }


\subsubsection{Evaluation Metrics}
\qy{Different from the evaluation of single-object tracking, performance evaluation for MOT is not straightforward due to the complexity of this task. To this end, we followed these evaluation metrics in generic multiple-object tracking \cite{TC,CLEAR}, and quantitatively evaluated the performance of these baselines on our \nickname{} dataset from the following three aspects:}

\qy{\textbf{Accuracy.} Multiple Object Tracking Accuracy (MOTA) is the most widely used metric to evaluate the overall performance of a multiple-object tracker. This metric is a combination of false positive rate, missed targets, and identity switches:}

\begin{equation}
\mathrm{MOTA} = 1- \frac{\sum_{t}(m_t+fp_t+mme_t)}{\sum_{t}{gt}_n}
\end{equation}

\noindent \qy{where $t$ is the frame index and $gt$ is the number of ground-truth objects. $m_t$, $fp_t$, and $mme_t$ denote the missed targets, false-positives, and ID switches in the $t$-th frame, respectively. MOTA describes the statistics of the accumulated errors in tracking, this metric ranges from (-$\infty$,100] in our baseline. Note that, a negative MOTA occurs when the number of errors made by the tracker is larger than the number of all objects in a video.}

\qy{\textbf{Precision.} These metrics are mainly used to measure how precisely the objects of interest are being tracked. It is obtained by calculating the bounding box overlap and/or center location distance. The Multiple Object Tracking Precision (MOTP) is a representative indicator to evaluate the precision performance of a tracker:}
\begin{equation}
\mathrm{MOTP} =  \frac{\sum_{t,i}d_{t}^{i}}{\sum_{t}c_t}
\end{equation}
\qy{where $c_t$ is the total number of matches made between true targets and hypothesized objects in frame $t$, and $d_{t}^{i}$ is the distance between the object $o_i$ and its corresponding hypothesis. The range of the MOTP indicator can be [0,100].} 

\qy{\textbf{Completeness.}  These metrics are used to evaluate the completeness of the entire tracking trajectory. In particular, the trajectory generated by grouping the tracker outputs can be classified as mostly tracked (MT), partially tracked (PT), and mostly lost (ML). The MT indicator means more than 80\% of the length of the ground-truth trajectory is covered by the tracker output, while ML means that less than 20\% of the length of the ground-truth trajectory is covered by the tracker output. All other cases are classified as PL. Therefore, an ideal tracker is expected to have a higher number of sequences classified as MT and a smaller number of ML.}

\qy{In addition, we also utilized several commonly used metrics for performance evaluation, including False Positive (FP), False Negative (FN), FM, and IDs. FM is the total number of a ground truth trajectory that is interrupted (untracked). IDs are defined as the total number of identity switches.}




\subsubsection{Experimental Analyses}


\qy{Quantitative results achieved by several baseline trackers are shown in Table \ref{Table_mot_results}. Note that, for trackers which do not have the detection module (\textit{e.g.}, Kalman Filter \cite{KF} and SORT \cite{Sort}), we use the detection results obtained by our detector as the input. For trackers such as D\&T \cite{DT}, we simply follow their framework and take the detection results as input. It can be seen that SORT \cite{Sort} achieves the best performance in terms of MOTP (\textit{i.e.}, 28.6\%). Kalman Filter \cite{KF} achieves the best performance in terms of MOTA (\textit{i.e.}, 73.6\%). However, we are also aware that the performance of all these baselines is far from satisfactory, where the majority of sequences are considered as mostly lost. This implies that existing MOT trackers cannot be well generalized to our dataset (\textit{i.e.}, satellite video sequences with a low resolution and insufficient information). }


\begin{table*}
\centering
\renewcommand\arraystretch{1.2}
\footnotesize
\setlength{\tabcolsep}{4.2mm}
\caption{Quantitative results achieved by several baselines in our \nickname. The up arrow (\textit{resp.} down arrow) indicates that the performance is better if the quantity is greater (\textit{resp.} smaller). The best results are in \textcolor{red}{\textbf{red}} and the second best results are in \textcolor{blue}{\textbf{blue}}. }\label{Table_mot_results}
\begin{tabular}{rccccccccccc}
\toprule[1.0pt]
Method & MOTA $\uparrow$ & MOTP$\uparrow$  & MT$\uparrow$  & PT$\downarrow$   & ML$\downarrow$  & FP$\downarrow$  & FN$\downarrow$ & IDS$\downarrow$ & FM$\downarrow$& FPS$\uparrow$ \tabularnewline
\toprule[1.0pt]
CMOT \cite{TC}& 22.8\% & 9.5\% & 38 & 111 & 494 & \textcolor{red}{0} & 71,638 & \textcolor{blue}{89} &\textcolor{red}{111}&17.2\\
FairMOT\cite{FairMOT} & 2.3\% & \textcolor{blue}{28.0\%} & 21 & \textcolor{red}{13} & 623 & 2,073 & 83,258 & \textcolor{red}{52}&\textcolor{blue}{205} & 19.1\\
DTTP\cite{Ahmadi2019}&44.5\%	&16.3\%	&483	&153	& \textcolor{blue}{22}	&38,329	&\textcolor{blue}{10,032}	&3,090	&1,344	&18.2\\
D\&T\cite{DT}& \textcolor{blue}{68.0\%}	&15.2\%	& \textcolor{blue}{511}	&71	&73	&15,862	&11,681	&2,122	&843&	24.7\\
Kalman \cite{KF} & \textcolor{red}{73.6\%} & 21.8\% & \textcolor{red}{639} &  \textcolor{blue}{19} & \textcolor{red}{0} & 21,589 & \textcolor{red}{774} & 2,085 &625 &  \textcolor{blue}{24.7}\\
SORT\cite{Sort} & 58.2\% & \textcolor{red}{28.6\%} & 214 & 218 & 221 & \textcolor{blue}{117} & 36,377 & 2,275 &2,047 &\textcolor{red}{92.9}\\
\toprule[1.0pt]
\end{tabular}

\end{table*}

\begin{figure}[t]
\centering
\includegraphics[width=9cm]{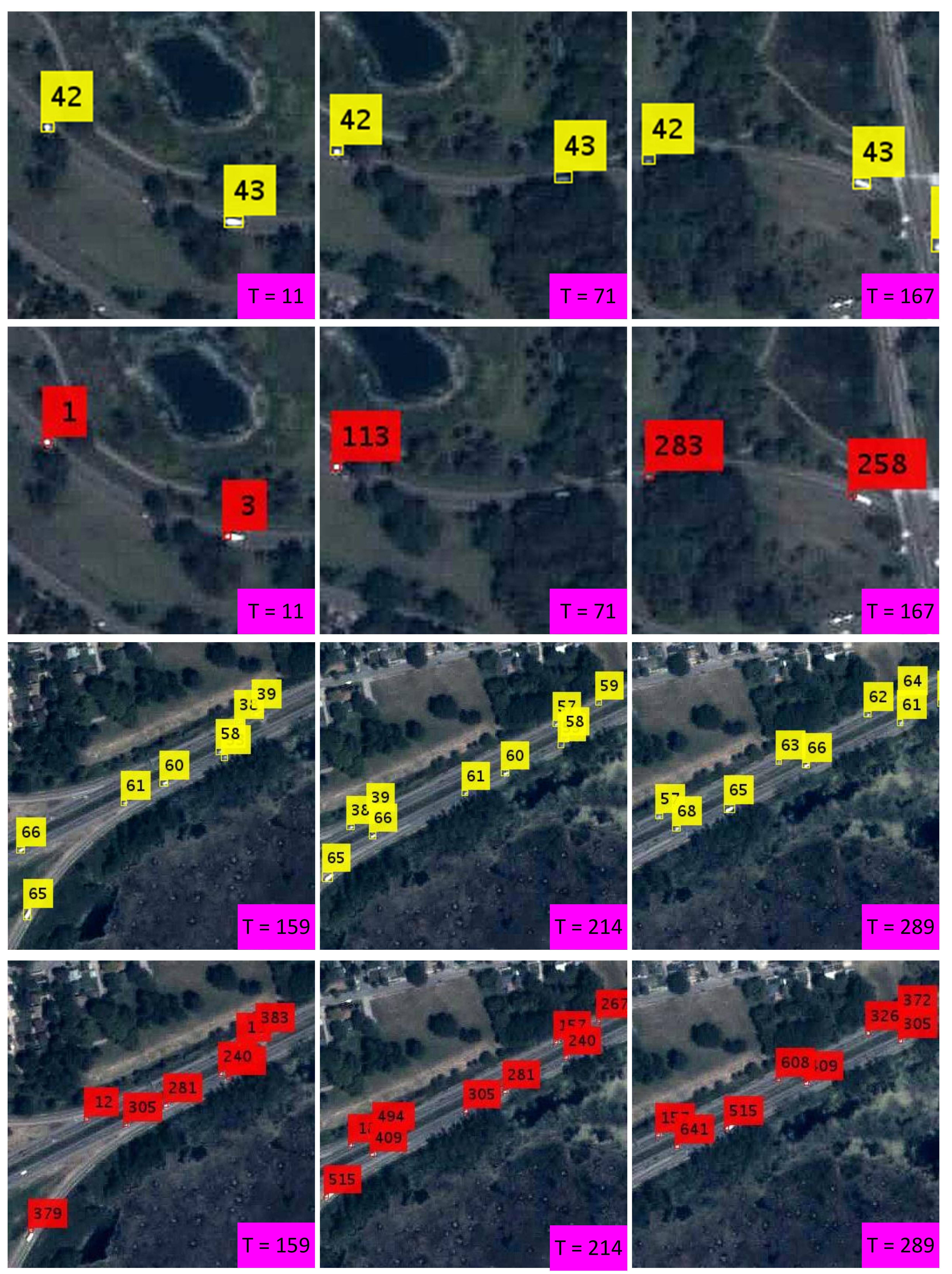}
\caption{Visualization of our multiple object tracking results on the VISO dataset. Note that, detection results are produced by our object detection method. \label{mot_vis}}
\end{figure}

\qy{To further investigate the reason for unsatisfactory tracking performance, we qualitatively show the results achieved on several typical scenarios on our \nickname{} dataset in Fig.~\ref{mot_vis}. It can be seen that the tracked objects in satellite video are usually small, but with similar appearance and texture. This poses a great challenge to baseline trackers. We can see that the ID switches of objects  occur frequently when a large number of objects are densely packed, which fundamentally degrades the overall tracking performance. In addition, occlusion and illumination variations make our dataset more challenging.}

\qy{In summary, it is still highly challenging to achieve multi-object tracking in satellite videos, especially when the number of objects is large and the resolution of objects is low. Existing MOT trackers are initially designed for generic videos and are unable to be directly generalized to satellite videos. As the first satellite MOT benchmarks, we believe our \nickname{} dataset will foster the research in this area.}

\section{Discussion}\label{discussion}
The proposed VISO dataset contains 47 long sequences, while each sequence contains an average of 325 frames captured in 32 seconds, which makes it quite suitable for the evaluation of persistent tracking algorithms \cite{pertracking}. This task requires the method being able to track the objects of interest once they start to move, and keep the tracking process as long as they are visible. Frame-differencing and background modeling methods are less effective in this setting, since they are unable to detect and track those temporary static objects. We encourage researchers to further investigate this research problem on our dataset.

\section{Conclusion}\label{conclusion}
In this paper, we introduced an urban-scale satellite video dataset for moving object detection and tracking. This dataset consists of 47 annotated videos captured by the Jilin-1 satellite constellation. This is a comprehensive  satellite video dataset with a number of unique challenges and functions. In addition, we also established a benchmark and evaluated several representative object detectors and trackers. We believe the proposed \nickname{} dataset and benchmark can advance the research in this community.
Besides, we proposed a moving object detector using both frame differencing and background subtraction. Extensive experimental results on the \nickname{} dataset show that the proposed method achieves high
detection performance and low false alarms.
In our future work, we would explore the possibility of expanding our VISO dataset to video segmentation and other relevant tasks.

\section*{Acknowledgment}
This work was partially supported by the National Natural Science Foundation of China (No. 61972435, U20A20185, 61872379), the Natural Science Foundation of Guangdong Province (2019A1515011271), the Science and Technology Innovation Committee of Shenzhen Municipality (JCYJ20190807152209394, RCYX20200714114641140), the China Scholarship Council (CSC).

\bibliographystyle{IEEEtran}
\bibliography{IEEEexample}


\ifCLASSOPTIONcaptionsoff
  \newpage
\fi





\begin{IEEEbiography}[{\includegraphics[width=1in,height=1.25in,clip,keepaspectratio]{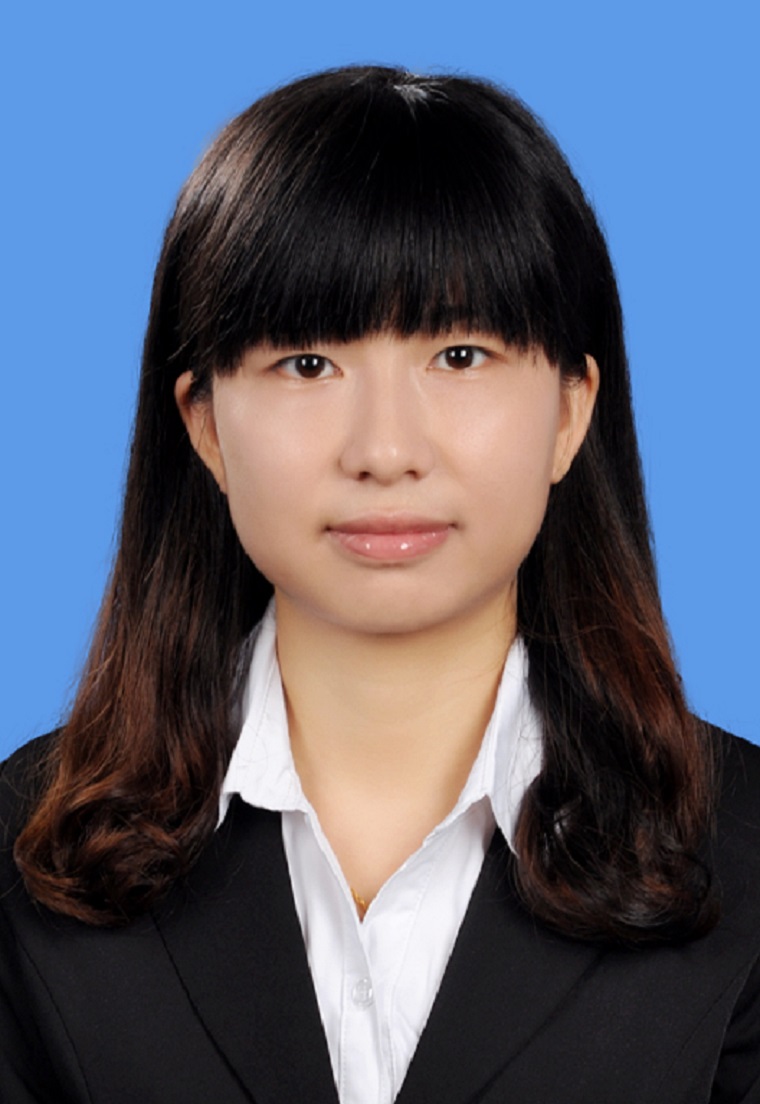}}]{Qian Yin} received the B.Eng. degree in electrical engineering from Hunan University of Science and Technology (HNUST), Xiangtan, China, in 2014, and the M.E. degree in optical engineering from South China Normal University (SCNU), Guangzhou, China, in 2016. She is currently pursuing the Ph.D. degree with the College of Electronic Science and Technology, National University of Defense Technology (NUDT). She research interests focus on object detection and tracking.
\end{IEEEbiography}

\begin{IEEEbiography}[{\includegraphics[width=1in,height=1.25in,clip,keepaspectratio]{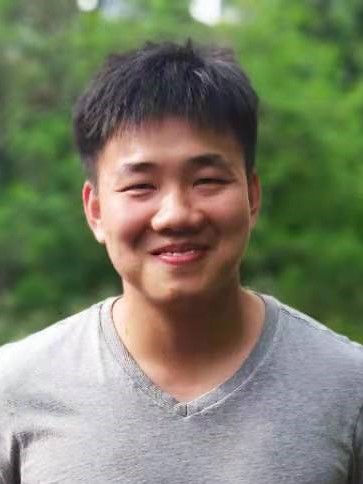}}]{Qingyong Hu} received his M.Eng. degree in information and communication engineering from the National University of Defense Technology (NUDT) in 2018. He is currently a DPhil candidate in the Department of Computer Science at the University of Oxford. His research interests lie in 3D computer vision, large-scale point cloud processing, and visual tracking.
\end{IEEEbiography}

\begin{IEEEbiography}[{\includegraphics[width=1in,height=1.25in,clip,keepaspectratio]{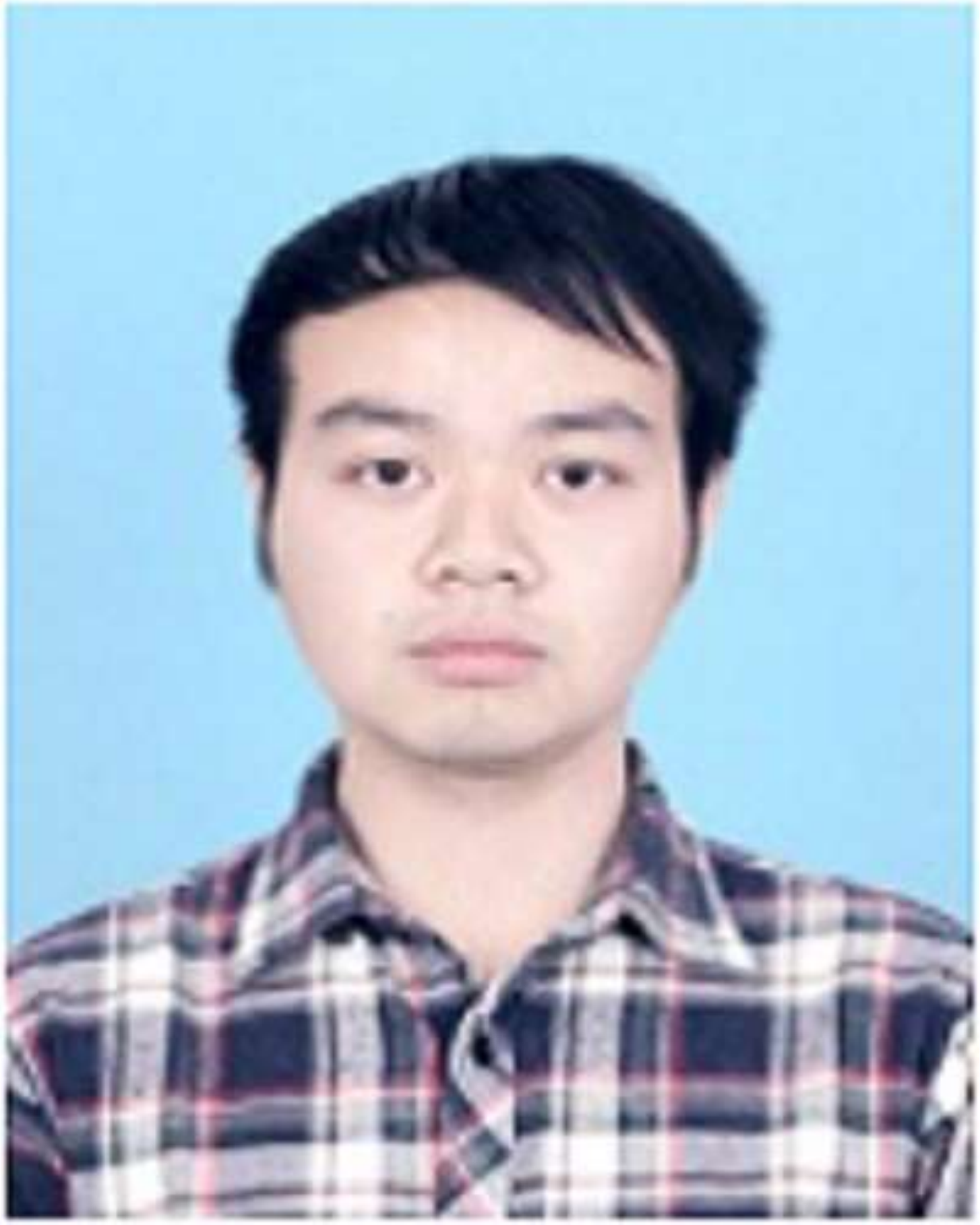}}]{Hao Liu} received the B.Eng. degree from University of Electronic Science and Technology of China (UESTC) in 2016, and M.S. degree from National University of Defense Technology (NUDT) in 2018. He is currently pursuing the Ph.D. degree with the School of Electronics and Communication Engineering, Sun Yat-sen University. His research interests lie in 3D computer vision and point cloud processing.
\end{IEEEbiography}
\begin{IEEEbiography}[{\includegraphics[width=1in,height=1.25in,clip,keepaspectratio]{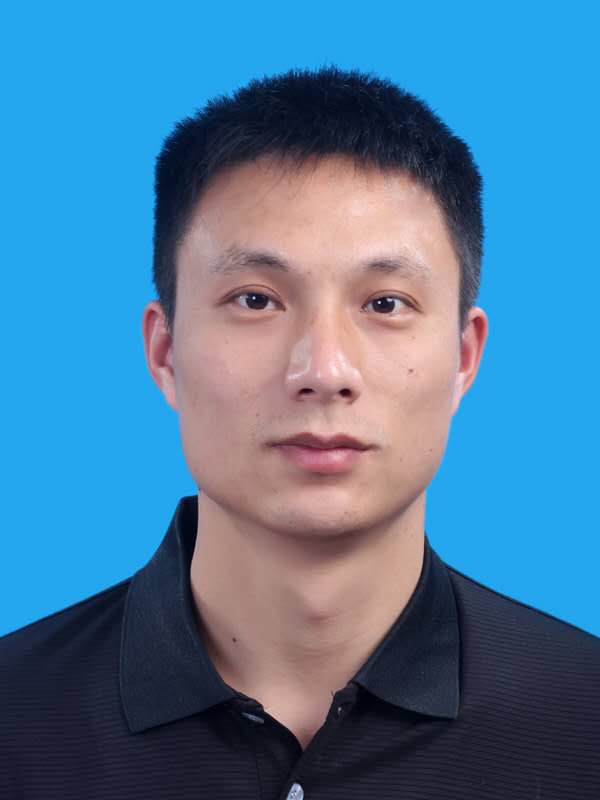}}]{Feng Zhang} receive the B.E degree in school of Electrics and Information Engineering Harbin Institute of Technology (HIT), Harbin, China, in 2009, and the M.E degree in College of Electronic Science and Technology, National University of Defense Technology (NUDT), Changsha, China, in 2011. He is currently purchase his Ph.D degree with the College of Electronic Science and Technology, NUDT. His main research interests focus on object detection and tracking.
\end{IEEEbiography}
\begin{IEEEbiography}[{\includegraphics[width=1in,height=1.25in,clip,keepaspectratio]{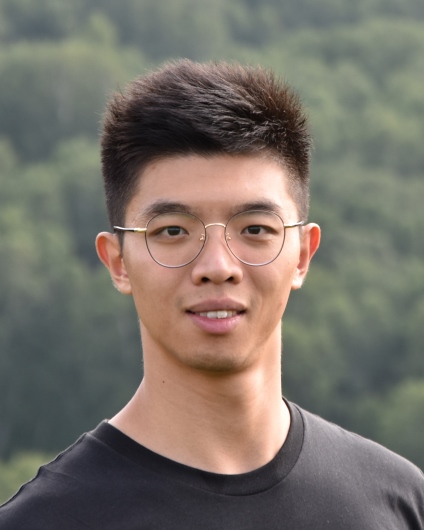}}]{Yingqian Wang} received the B.E. degree in electrical engineering from Shandong University (SDU), Jinan, China, in 2016, and the M.E. degree in information and communication engineering from National University of Defense Technology (NUDT), Changsha, China, in 2018. He is currently pursuing the Ph.D. degree with the College of Electronic Science and Technology, NUDT. His research interests focus on low-level vision, particularly on light field imaging and image super-resolution.
\end{IEEEbiography}
\begin{IEEEbiography}[{\includegraphics[width=1in,height=1.25in,clip,keepaspectratio]{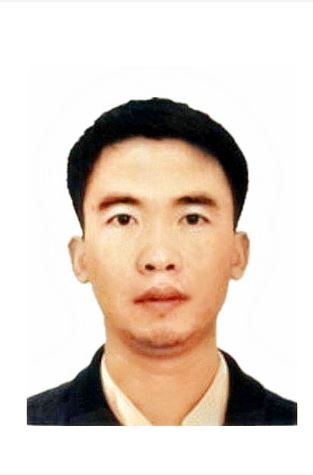}}]{Zaiping Lin} received the B.E. degree and Ph.D. degree from the National University of Defense Technology (NUDT), Changsha, China, in 2007 and 2012, respectively. He is currently
an Assistant Professor with the College Electronic Science and Technology, NUDT. His current research interests include infrared image processing and signal processing.
\end{IEEEbiography}
\begin{IEEEbiography}[{\includegraphics[width=1in,height=1.25in,clip,keepaspectratio]{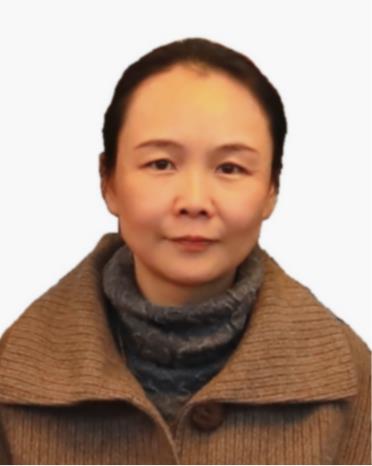}}]{Wei An} received the Ph.D. degree from the National University of Defense Technology (NUDT), Changsha, China, in 1999. She was a Senior Visiting Scholar with the University of Southampton, Southampton, U.K., in 2016. She is currently a Professor with the College of Electronic Science and Technology, NUDT. She has authored or co-authored over 100 journal and conference publications. Her current research interests include signal processing and image processing.
\end{IEEEbiography}

\begin{IEEEbiography}[{\includegraphics[width=1in,height=1.25in,clip,keepaspectratio]{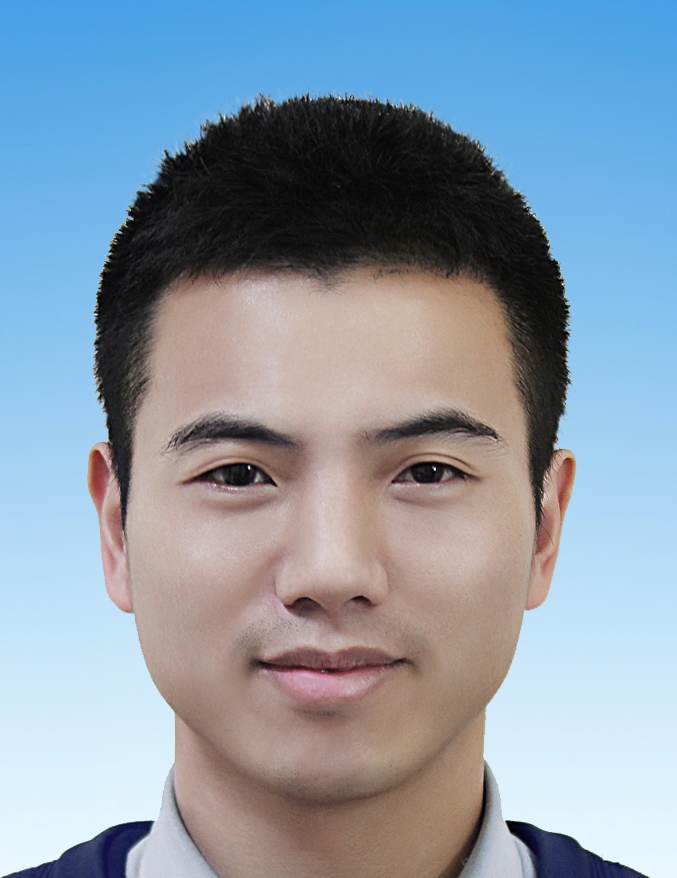}}]{Yulan Guo} is currently an associate professor. He received the B.Eng. and Ph.D. degrees from National University of Defense Technology (NUDT) in 2008 and 2015, respectively. He was a visiting Ph.D. student with the University of Western Australia from 2011 to 2014. He worked as a postdoctorial research fellow with the Institute of Computing Technology, Chinese Academy of Sciences from 2016 to 2018. He has authored over 100 articles in journals and conferences. 
His current research interests focus on 3D vision, particularly on 3D feature learning, 3D modeling, 3D object recognition, and scene understanding. Dr. Guo received the ACM China SIGAI Rising Star Award in 2019, Wu-Wenjun Outstanding AI Youth Award in 2019, and the CAAI Outstanding Doctoral Dissertation Award in 2016. He served as an associate editor for IET Computer Vision, IET Image Processing, and Computers \& Graphics, a guest editor for IEEE TPAMI, and an area chair for CVPR 2021, ICCV 2021, ACM Multimedia 2021, and ICPR 2020. He is a senior member of IEEE.
\end{IEEEbiography}
\end{document}